\documentclass{article}
\usepackage{enumitem}
\usepackage{cite}
\usepackage{amsmath,amssymb,amsfonts}
\usepackage{algorithmic}
\usepackage{graphicx}
\usepackage{textcomp}
\usepackage{tabularx}
\usepackage{xcolor}
\usepackage{listings}
\usepackage[frozencache,cachedir=.]{minted}

\usepackage{cleveref}
\crefname{requirementsi}{Requirement}{requirements}
\Crefname{requirementsi}{Requirement}{Requirements}

\usepackage{tikz}
\usepackage{ifthen}

\newcommand{\harveyball}[1]{%
  \begin{tikzpicture}[scale=0.14]
    \draw (0,0) circle (1); 
    \ifthenelse{\equal{#1}{0}}{}{%
      \ifthenelse{\equal{#1}{25}}{%
        \fill (0,0) -- (180:1) arc (180:270:1) -- cycle; 
      }{
        \ifthenelse{\equal{#1}{50}}{%
          \fill (0,0) -- (90:1) arc (90:270:1) -- cycle; 
        }{
          \ifthenelse{\equal{#1}{75}}{%
            \fill (0,0) -- (180:1) arc (180:450:1) -- cycle; 
          }{
            \ifthenelse{\equal{#1}{100}}{%
              \fill (0,0) circle (1); 
            }{}
          }
        }
      }
    }
  \end{tikzpicture}%
}

\makeatletter
\def\PYG@reset{\let\PYG@it=\relax \let\PYG@bf=\relax%
    \let\PYG@ul=\relax \let\PYG@tc=\relax%
    \let\PYG@bc=\relax \let\PYG@ff=\relax}
\def\PYG@tok#1{\csname PYG@tok@#1\endcsname}
\def\PYG@toks#1+{\ifx\relax#1\empty\else%
    \PYG@tok{#1}\expandafter\PYG@toks\fi}
\def\PYG@do#1{\PYG@bc{\PYG@tc{\PYG@ul{%
    \PYG@it{\PYG@bf{\PYG@ff{#1}}}}}}}
\def\PYG#1#2{\PYG@reset\PYG@toks#1+\relax+\PYG@do{#2}}

\@namedef{PYG@tok@w}{\def\PYG@tc##1{\textcolor[rgb]{0.73,0.73,0.73}{##1}}}
\@namedef{PYG@tok@c}{\let\PYG@it=\textit\def\PYG@tc##1{\textcolor[rgb]{0.24,0.48,0.48}{##1}}}
\@namedef{PYG@tok@cp}{\def\PYG@tc##1{\textcolor[rgb]{0.61,0.40,0.00}{##1}}}
\@namedef{PYG@tok@k}{\let\PYG@bf=\textbf\def\PYG@tc##1{\textcolor[rgb]{0.00,0.50,0.00}{##1}}}
\@namedef{PYG@tok@kp}{\def\PYG@tc##1{\textcolor[rgb]{0.00,0.50,0.00}{##1}}}
\@namedef{PYG@tok@kt}{\def\PYG@tc##1{\textcolor[rgb]{0.69,0.00,0.25}{##1}}}
\@namedef{PYG@tok@o}{\def\PYG@tc##1{\textcolor[rgb]{0.40,0.40,0.40}{##1}}}
\@namedef{PYG@tok@ow}{\let\PYG@bf=\textbf\def\PYG@tc##1{\textcolor[rgb]{0.67,0.13,1.00}{##1}}}
\@namedef{PYG@tok@nb}{\def\PYG@tc##1{\textcolor[rgb]{0.00,0.50,0.00}{##1}}}
\@namedef{PYG@tok@nf}{\def\PYG@tc##1{\textcolor[rgb]{0.00,0.00,1.00}{##1}}}
\@namedef{PYG@tok@nc}{\let\PYG@bf=\textbf\def\PYG@tc##1{\textcolor[rgb]{0.00,0.00,1.00}{##1}}}
\@namedef{PYG@tok@nn}{\let\PYG@bf=\textbf\def\PYG@tc##1{\textcolor[rgb]{0.00,0.00,1.00}{##1}}}
\@namedef{PYG@tok@ne}{\let\PYG@bf=\textbf\def\PYG@tc##1{\textcolor[rgb]{0.80,0.25,0.22}{##1}}}
\@namedef{PYG@tok@nv}{\def\PYG@tc##1{\textcolor[rgb]{0.10,0.09,0.49}{##1}}}
\@namedef{PYG@tok@no}{\def\PYG@tc##1{\textcolor[rgb]{0.53,0.00,0.00}{##1}}}
\@namedef{PYG@tok@nl}{\def\PYG@tc##1{\textcolor[rgb]{0.46,0.46,0.00}{##1}}}
\@namedef{PYG@tok@ni}{\let\PYG@bf=\textbf\def\PYG@tc##1{\textcolor[rgb]{0.44,0.44,0.44}{##1}}}
\@namedef{PYG@tok@na}{\def\PYG@tc##1{\textcolor[rgb]{0.41,0.47,0.13}{##1}}}
\@namedef{PYG@tok@nt}{\let\PYG@bf=\textbf\def\PYG@tc##1{\textcolor[rgb]{0.00,0.50,0.00}{##1}}}
\@namedef{PYG@tok@nd}{\def\PYG@tc##1{\textcolor[rgb]{0.67,0.13,1.00}{##1}}}
\@namedef{PYG@tok@s}{\def\PYG@tc##1{\textcolor[rgb]{0.73,0.13,0.13}{##1}}}
\@namedef{PYG@tok@sd}{\let\PYG@it=\textit\def\PYG@tc##1{\textcolor[rgb]{0.73,0.13,0.13}{##1}}}
\@namedef{PYG@tok@si}{\let\PYG@bf=\textbf\def\PYG@tc##1{\textcolor[rgb]{0.64,0.35,0.47}{##1}}}
\@namedef{PYG@tok@se}{\let\PYG@bf=\textbf\def\PYG@tc##1{\textcolor[rgb]{0.67,0.36,0.12}{##1}}}
\@namedef{PYG@tok@sr}{\def\PYG@tc##1{\textcolor[rgb]{0.64,0.35,0.47}{##1}}}
\@namedef{PYG@tok@ss}{\def\PYG@tc##1{\textcolor[rgb]{0.10,0.09,0.49}{##1}}}
\@namedef{PYG@tok@sx}{\def\PYG@tc##1{\textcolor[rgb]{0.00,0.50,0.00}{##1}}}
\@namedef{PYG@tok@m}{\def\PYG@tc##1{\textcolor[rgb]{0.40,0.40,0.40}{##1}}}
\@namedef{PYG@tok@gh}{\let\PYG@bf=\textbf\def\PYG@tc##1{\textcolor[rgb]{0.00,0.00,0.50}{##1}}}
\@namedef{PYG@tok@gu}{\let\PYG@bf=\textbf\def\PYG@tc##1{\textcolor[rgb]{0.50,0.00,0.50}{##1}}}
\@namedef{PYG@tok@gd}{\def\PYG@tc##1{\textcolor[rgb]{0.63,0.00,0.00}{##1}}}
\@namedef{PYG@tok@gi}{\def\PYG@tc##1{\textcolor[rgb]{0.00,0.52,0.00}{##1}}}
\@namedef{PYG@tok@gr}{\def\PYG@tc##1{\textcolor[rgb]{0.89,0.00,0.00}{##1}}}
\@namedef{PYG@tok@ge}{\let\PYG@it=\textit}
\@namedef{PYG@tok@gs}{\let\PYG@bf=\textbf}
\@namedef{PYG@tok@gp}{\let\PYG@bf=\textbf\def\PYG@tc##1{\textcolor[rgb]{0.00,0.00,0.50}{##1}}}
\@namedef{PYG@tok@go}{\def\PYG@tc##1{\textcolor[rgb]{0.44,0.44,0.44}{##1}}}
\@namedef{PYG@tok@gt}{\def\PYG@tc##1{\textcolor[rgb]{0.00,0.27,0.87}{##1}}}
\@namedef{PYG@tok@err}{\def\PYG@bc##1{{\setlength{\fboxsep}{\string -\fboxrule}\fcolorbox[rgb]{1.00,0.00,0.00}{1,1,1}{\strut ##1}}}}
\@namedef{PYG@tok@kc}{\let\PYG@bf=\textbf\def\PYG@tc##1{\textcolor[rgb]{0.00,0.50,0.00}{##1}}}
\@namedef{PYG@tok@kd}{\let\PYG@bf=\textbf\def\PYG@tc##1{\textcolor[rgb]{0.00,0.50,0.00}{##1}}}
\@namedef{PYG@tok@kn}{\let\PYG@bf=\textbf\def\PYG@tc##1{\textcolor[rgb]{0.00,0.50,0.00}{##1}}}
\@namedef{PYG@tok@kr}{\let\PYG@bf=\textbf\def\PYG@tc##1{\textcolor[rgb]{0.00,0.50,0.00}{##1}}}
\@namedef{PYG@tok@bp}{\def\PYG@tc##1{\textcolor[rgb]{0.00,0.50,0.00}{##1}}}
\@namedef{PYG@tok@fm}{\def\PYG@tc##1{\textcolor[rgb]{0.00,0.00,1.00}{##1}}}
\@namedef{PYG@tok@vc}{\def\PYG@tc##1{\textcolor[rgb]{0.10,0.09,0.49}{##1}}}
\@namedef{PYG@tok@vg}{\def\PYG@tc##1{\textcolor[rgb]{0.10,0.09,0.49}{##1}}}
\@namedef{PYG@tok@vi}{\def\PYG@tc##1{\textcolor[rgb]{0.10,0.09,0.49}{##1}}}
\@namedef{PYG@tok@vm}{\def\PYG@tc##1{\textcolor[rgb]{0.10,0.09,0.49}{##1}}}
\@namedef{PYG@tok@sa}{\def\PYG@tc##1{\textcolor[rgb]{0.73,0.13,0.13}{##1}}}
\@namedef{PYG@tok@sb}{\def\PYG@tc##1{\textcolor[rgb]{0.73,0.13,0.13}{##1}}}
\@namedef{PYG@tok@sc}{\def\PYG@tc##1{\textcolor[rgb]{0.73,0.13,0.13}{##1}}}
\@namedef{PYG@tok@dl}{\def\PYG@tc##1{\textcolor[rgb]{0.73,0.13,0.13}{##1}}}
\@namedef{PYG@tok@s2}{\def\PYG@tc##1{\textcolor[rgb]{0.73,0.13,0.13}{##1}}}
\@namedef{PYG@tok@sh}{\def\PYG@tc##1{\textcolor[rgb]{0.73,0.13,0.13}{##1}}}
\@namedef{PYG@tok@s1}{\def\PYG@tc##1{\textcolor[rgb]{0.73,0.13,0.13}{##1}}}
\@namedef{PYG@tok@mb}{\def\PYG@tc##1{\textcolor[rgb]{0.40,0.40,0.40}{##1}}}
\@namedef{PYG@tok@mf}{\def\PYG@tc##1{\textcolor[rgb]{0.40,0.40,0.40}{##1}}}
\@namedef{PYG@tok@mh}{\def\PYG@tc##1{\textcolor[rgb]{0.40,0.40,0.40}{##1}}}
\@namedef{PYG@tok@mi}{\def\PYG@tc##1{\textcolor[rgb]{0.40,0.40,0.40}{##1}}}
\@namedef{PYG@tok@il}{\def\PYG@tc##1{\textcolor[rgb]{0.40,0.40,0.40}{##1}}}
\@namedef{PYG@tok@mo}{\def\PYG@tc##1{\textcolor[rgb]{0.40,0.40,0.40}{##1}}}
\@namedef{PYG@tok@ch}{\let\PYG@it=\textit\def\PYG@tc##1{\textcolor[rgb]{0.24,0.48,0.48}{##1}}}
\@namedef{PYG@tok@cm}{\let\PYG@it=\textit\def\PYG@tc##1{\textcolor[rgb]{0.24,0.48,0.48}{##1}}}
\@namedef{PYG@tok@cpf}{\let\PYG@it=\textit\def\PYG@tc##1{\textcolor[rgb]{0.24,0.48,0.48}{##1}}}
\@namedef{PYG@tok@c1}{\let\PYG@it=\textit\def\PYG@tc##1{\textcolor[rgb]{0.24,0.48,0.48}{##1}}}
\@namedef{PYG@tok@cs}{\let\PYG@it=\textit\def\PYG@tc##1{\textcolor[rgb]{0.24,0.48,0.48}{##1}}}

\def\PYGZus{\char`\_}
\def\PYGZob{\char`\{}
\def\PYGZcb{\char`\}}

\def\PYGZlt{\char`\<}
\def\PYGZgt{\char`\>}

\def\PYGZdl{\char`\$}
\def\PYGZhy{\char`\-}
\def\PYGZsq{\char`\'}
\def\PYGZdq{\char`\"}

\makeatother

\newminted[AutomationML]{xml}{
    frame=lines,
    bgcolor=gray!10,
    fontsize=\footnotesize,
    tabsize=2
}
\newminted[SPARQL]{sparql}{
    frame=lines,
    bgcolor=gray!10,
    fontsize=\footnotesize,
    tabsize=2
}
\newminted[xPath]{xpath}{
    frame=lines,
    bgcolor=gray!10,
    fontsize=\footnotesize,
    tabsize=2
}
\newminted[xQuery]{xquery}{
    frame=lines,
    bgcolor=gray!10,
    fontsize=\footnotesize,
    tabsize=2
}

\def\BibTeX{{\rm B\kern-.05em{\sc i\kern-.025em b}\kern-.08em
    T\kern-.1667em\lower.7ex\hbox{E}\kern-.125emX}}
\title{Automatic Mapping of AutomationML Files to Ontologies for Graph Queries and Validation}
\begin{document}
\author{
  Tom Westermann$^{1}$\thanks{Corresponding author: \texttt{tom.westermann@hsu-hh.de}},
  Malte Ramonat$^{1}$,
  Johannes Hujer$^{1}$,\\
  Felix Gehlhoff$^{1}$,
  Alexander Fay$^{2}$ 
  \\
  \\
  $^1$Institute of Automation Technology, \\Helmut Schmidt University, Hamburg, Germany \\
  $^2$Chair of Automation Technology, \\Ruhr University, Bochum, Germany
}

\date{}
\maketitle

\renewcommand{\thefootnote}{\fnsymbol{footnote}}
\footnotetext[1]{This research article [project ProMoDi] is funded by dtec.bw – Digitalization and Technology Research Center of the Bundeswehr. dtec.bw is funded by the European Union – NextGenerationEU.\\
This research is also funded by the Federal Ministry for Economic Affairs and Climate Action (BMWK) on the basis of a decision by the German Bundestag.}


\begin{abstract}
AutomationML has seen widespread adoption as an open data exchange format in the automation domain. 
It is an open and vendor neutral standard based on the extensible markup language XML. 
However, AutomationML extends XML with additional semantics that limit the applicability of common XML-tools for applications like querying or data validation. 
This article demonstrates how the transformation of AutomationML into OWL enables new use cases in querying with SPARQL and validation with SHACL. To support this, it provides practitioners with (1) an up-to-date ontology of the concepts defined in the AutomationML standard and (2) a declarative mapping to automatically transform any AutomationML model into RDF triples.
A study on examples from the automation domain concludes that transforming AutomationML to OWL opens up new powerful ways for querying and validation that would have been impossible without this transformation. 
\end{abstract}


\maketitle

\section{Introduction}\label{sec:introduction}
AutomationML is an open and vendor neutral solution for data exchange in the domain of automation engineering. 
Since its first publication in 2008~\cite{Drath.2008}, it has seen widespread adoption in both the scientific community as well as industry, where it is viewed as a central building block for digital twins \cite{Talkhestani.2019}.

AutomationML is designed to represent complex relationships within engineering projects. 
It is used to model and exchange data related to the design, implementation, and maintenance of automation systems.
The AutomationML data model takes the form of a deeply interconnected graph. 
However, for easier exchange with other engineering tools, AutomationML uses XML as its underlying format for serialization, which follows a tree structure. In the serialization, references between AutomationML elements of the XML-tree are realized through reference paths and identifiers. 

In practice, this representation mechanism offers multiple drawbacks. 
First, commonly used tools to query and validate XML-Files (like XSD or xQuery) were designed to operate on the hierarchical structure of the data format. They therefore have very rudimentary support for operations involving the additional connections defined by AutomationML. 
Second, the AutomationML data lacks a formal representation which would allow knowledge inference and further analytical applications\cite{OlgaKovalenko.2018}.
Third, the XML-Schema of AutomationML does not allow for native alignment with other data sources, i.e. it is limited to the vocabulary that AutomationML has to offer.
However, this alignment is a pillar of the digital twin\cite{Talkhestani.2019}, making it especially necessary in the engineering domain \cite{Jirkovsky.2017}.

In terms of formally representing engineering knowledge, technologies like the Resource Description Framework (RDF) and the Web Ontology Language (OWL) are widely used, as they provide a standardized framework for expressing relationships and semantics \cite{Kalibatien2011}.
These technologies have been applied to create meta models of cyber-physical systems \cite{Derler.2012}\cite{Hildebrandt.2020}, which allows for the integration of heterogeneous data and resolving semantic heterogeneity \cite{Jirkovsky.2017}. RDF and OWL further allow the use of technologies from the Semantic Web Stack like SPARQL and SHACL, which further enable efficient querying and validation of the created knowledge graphs.

The transformation of AutomationML into an ontology-compliant knowledge graph has been in the focus of the research community for a while\cite{L.Abele.2015, GrangelGonzalez.2016, GrangelGonzalez.2018, OlgaKovalenko.2018, Goncalves.2019}, usually with a focus on data integration. 

However, none of the existing approaches have systematically examined the extent to which such a transformation enables new applications. While prior work has primarily emphasized data integration, the potential benefits of representing AutomationML as an ontology—such as the ability to formulate expressive semantic queries or to apply logical validation techniques—remain largely unexplored.

To address this gap, the paper compares the querying capabilities of XML-based tools with those of SPARQL, and examines how validation in XML-based environments contrasts with the possibilities offered by SHACL. In doing so, it introduces an up-to-date AutomationML ontology that covers the full vocabulary of the standard, presents a declarative mapping that enables the straightforward transformation of AutomationML files into RDF triples and provides a software implementation of this mapping to support its practical application. Together, these contributions provide practitioners with both the technical foundations and practical means to exploit ontology-based querying and validation in AutomationML beyond the limits of existing XML-based approaches.

The remainder of the article is structured as follows: 
Section~\ref{sec:Background} gives more information on the AutomationML format, how it relates to graph theory and the tools from the Semantic Web Stack. 
Section~\ref{sec:RequirementsAndSOTA} describes previous attempts to create an AutomationML ontology and a mapping, while Section~\ref{sec:AutomationMLontology} presents the newly proposed ontology. 
Section~\ref{sec:RmlMapping} formalizes the mapping rules to transform AutomationML files to OWL, which are exemplified in Section \ref{sec:UseCase}.  
Section~\ref{sec:Usage} evaluates the approach by comparing querying with XML-based tools and SPARQL, as well as validation with XML-based mechanisms and SHACL.
The use cases show that transforming AutomationML to RDF allows for more complex graph queries and validation rules in comparison to typical XML-based approaches. 
\section{Background}\label{sec:Background}

\subsection{AutomationML}\label{sec:AutomationML}
AutomationML (Automation Markup Language) is an open, XML-based standard for the exchange of engineering data.
It builds on established and standardized data exchange formats like Computer Aided Engineering Exchange (CAEX) for plant topology\cite{IEC62424-2016}, COLLADA for geometry and kinematics \cite{ISO17506-2022}, and PLCopenXML for control logic\cite{IEC61131-10-2019}.  
It is commonly used in both the scientific community and industry\cite{Drath.2008, Mersch.2023} and has subsequently been standardized in IEC 62714\cite{IEC62714-1:2018}. 

\begin{figure}[htbp]
\includegraphics[width=\linewidth]{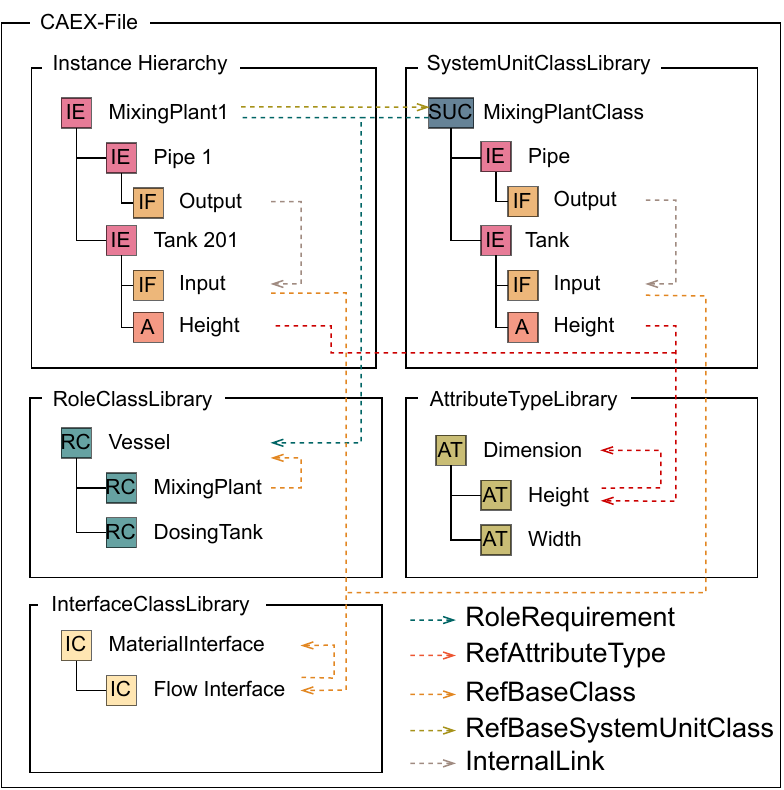}
\caption{The five pillars of CAEX in an AutomationML model alongside some of the most common reference types.}
\label{fig:automationMLFile}
\end{figure}

Recently, a new normative document - the IEC 62714-2\cite{IEC62714-2:2022} -  was released, which expands the scope of the AutomationML standard by introducing semantics libraries.
Including the newly added AttributeTypeLibrary, the CAEX submodel of AutomationML now consists of five main pillars (see Fig.~\ref{fig:automationMLFile})\cite{Drath.2021}:
\begin{itemize}
    \item The \textit{InstanceHierarchy} is a hierarchy of InternalElements (IE) and contains the data of a specific project. 
    \item The \textit{SystemUnitClassLibrary} provides SystemUnitClasses (SUC), which describe types of physical plant objects, e.g. a specific type of robot or valve. 
    \item The \textit{RoleClassLibrary} provides RoleClasses (RC), which define abstract roles that a physical or logical plant objects may fill. While SystemUnitClasses usually describe some manufacturer specific systems, RoleClasses model the abstract functions of components. 
    \item The \textit{InterfaceClassLibrary} provides InterfaceClasses (IC) which define a type of interface (e.g. a flange or a signal interface). These interfaces are used to describe physical or logical connections between InternalElements (IE) and SystemUnitClasses (SUC) through InternalLinks. 
    \item The \textit{AttributeTypeLibrary} defines a predefined data structure of reusable AttributeTypes (AT) and can be used to model attributes with values, units and descriptions. It allows for the definition of proprietary or standardized vocabularies of attributes, and creates a basis for the semantic interpretability of attributes. 
\end{itemize}

The content of an AutomationML file is structured in the shape of an XML-hierarchy. However, the semantics of this hierarchy can vary depending on the context within the AutomationML structure. For example, in the instance hierarchy, parent-child relationships represent physical or logical containment. 
In contrast, within class libraries such as RoleClassLibraries, AttributeTypeLibraries and InterfaceClassLibraries, the hierarchy is primarily used for structuring and organizing the data. Notably, it does not imply inheritance or specialization by default, but rather provides a logical grouping of related elements for better readability and maintainability.

Beyond the hierarchical organization, AutomationML supports multiple referencing mechanisms to relate elements across the file. Most references -- such as those used for class inheritance, interface instantiation, and role or class assignments -- are based on a reference path system. These reference paths point to elements within the structured hierarchy using relative or absolute identifiers, allowing connections to be expressed in a way that reflects the modular and nested organization of the XML structure.

In contrast, InternalLinks are used to explicitly define connections between Interface elements by referencing their unique IDs. Since Interfaces typically represent defined interaction points—such as mechanical, electrical, or software interfaces—InternalLinks provide a way to model physical or logical connections between them. This mechanism enables the representation of cross-hierarchy associations that are not captured by structural containment alone. 

Listing \ref{list:AutomationMLReferenceTypes} shows examples of hierarchy-, path- and ID-based references. 

\begin{listing}[ht]
\caption{Excerpt from the InstanceHierarchy depicted in Figure~\ref{fig:automationMLFile} showing all three reference mechanisms: Hierarchy-, path-, and ID-based.}
\label{list:AutomationMLReferenceTypes}
\begin{Verbatim}[commandchars=\\\{\}]
\PYG{n+nt}{\PYGZlt{}InstanceHierarchy}\PYG{+w}{ }\PYG{n+na}{Name=}\PYG{l+s}{\PYGZdq{}MyInstanceHierarchy\PYGZdq{}}\PYG{n+nt}{\PYGZgt{}}
\PYG{+w}{ }\PYG{n+nt}{\PYGZlt{}InternalElement}\PYG{+w}{ }\PYG{n+na}{Name=}\PYG{l+s}{\PYGZdq{}MixingPlant1\PYGZdq{}}\PYG{n+nt}{\PYGZgt{}}
\PYG{+w}{  }\PYG{n+nt}{\PYGZlt{}InternalElement}\PYG{+w}{ }\PYG{n+na}{Name=}\PYG{l+s}{\PYGZdq{}Pipe1\PYGZdq{}}\PYG{n+nt}{\PYGZgt{}}
\PYG{+w}{   }\PYG{n+nt}{\PYGZlt{}ExternalInterface}\PYG{+w}{ }\PYG{n+na}{Name=}\PYG{l+s}{\PYGZdq{}Output\PYGZdq{}}
\PYG{+w}{    }\PYG{n+na}{ID=}\PYG{l+s}{\PYGZdq{}6eea7a40\PYGZhy{}43fd\PYGZhy{}4aee\PYGZhy{}a1d3\PYGZdq{}}\PYG{n+nt}{\PYGZgt{}}
\PYG{+w}{   }\PYG{n+nt}{\PYGZlt{}/ExternalInterface\PYGZgt{}}
\PYG{+w}{  }\PYG{n+nt}{\PYGZlt{}/InternalElement\PYGZgt{}}
\PYG{+w}{  }\PYG{n+nt}{\PYGZlt{}InternalElement}\PYG{+w}{ }\PYG{n+na}{Name=}\PYG{l+s}{\PYGZdq{}Tank1\PYGZdq{}}\PYG{n+nt}{\PYGZgt{}}
\PYG{+w}{   }\PYG{n+nt}{\PYGZlt{}ExternalInterface}\PYG{+w}{ }\PYG{n+na}{Name=}\PYG{l+s}{\PYGZdq{}Input\PYGZdq{}}
\PYG{+w}{    }\PYG{n+na}{ID=}\PYG{l+s}{\PYGZdq{}ce45d7b3\PYGZhy{}169d\PYGZhy{}4be8\PYGZhy{}9746\PYGZdq{}}\PYG{n+nt}{\PYGZgt{}}
\PYG{+w}{   }\PYG{n+nt}{\PYGZlt{}/ExternalInterface\PYGZgt{}}
\PYG{+w}{   }\PYG{n+nt}{\PYGZlt{}Attribute}\PYG{+w}{ }\PYG{n+na}{Name=}\PYG{l+s}{\PYGZdq{}Length\PYGZdq{}}
\PYG{+w}{    }\PYG{n+na}{RefAttributeType=}\PYG{l+s}{\PYGZdq{}MyAtLib/Dimensions/Length\PYGZdq{}}\PYG{n+nt}{\PYGZgt{}}
\PYG{+w}{   }\PYG{n+nt}{\PYGZlt{}/Attribute\PYGZgt{}}
\PYG{+w}{   }\PYG{n+nt}{\PYGZlt{}InternalLink}\PYG{+w}{ }\PYG{n+na}{Name=}\PYG{l+s}{\PYGZdq{}Pipe1\PYGZus{}to\PYGZus{}Tank1\PYGZdq{}}
\PYG{+w}{    }\PYG{n+na}{RefPartnerSideA=}\PYG{l+s}{\PYGZdq{}6eea7a40\PYGZhy{}43fd\PYGZhy{}4aee\PYGZhy{}a1d3\PYGZdq{}}
\PYG{+w}{    }\PYG{n+na}{RefPartnerSideB=}\PYG{l+s}{\PYGZdq{}ce45d7b3\PYGZhy{}169d\PYGZhy{}4be8\PYGZhy{}9746\PYGZdq{}}\PYG{n+nt}{/\PYGZgt{}}
\PYG{+w}{   }\PYG{n+nt}{\PYGZlt{}/InternalElement\PYGZgt{}}
\PYG{+w}{  }\PYG{n+nt}{\PYGZlt{}/InternalElement\PYGZgt{}}
\PYG{n+nt}{\PYGZlt{}/InstanceHierarchy\PYGZgt{}}
\end{Verbatim}
\end{listing}   

AutomationML adopts the CAEX data model and terminology, but some terms acquire context-dependent meanings within AutomationML. Certain AutomationML specific concepts, such as \textit{Groups} or \textit{Facets}, are technically modeled as InternalElement objects with a predefined RoleRequirement, thereby specializing standard CAEX constructs. 
In other cases, semantics shift depending on the connection type—for example, a \textit{RefBaseSystemUnitClass} link may either indicate an association between an InternalElement and a SystemUnitClass, or represent a master/mirror relationship between two InternalElements.

\subsection{AutomationML and Graph Theory}\label{sec:GraphTheory}
In mathematical terms, a graph \( G \) is defined as an ordered pair:
\begin{equation}
G = (V, E)
\end{equation}
where:
\begin{itemize}\label{eq:Graph}
    \item[\( V \)] is a set of vertices (nodes), and
    \item[\( E \)] is a set of edges, where each edge is a pair of distinct vertices:
    \[ E \subseteq \left\{ \{u, v\} \mid u, v \in V, \; u \neq v \right\}\]
\end{itemize}
Graphs can take on various shapes and possess different properties, depending on how their vertices and edges are arranged.

\begin{figure}[tbp]
\includegraphics[width=\linewidth]{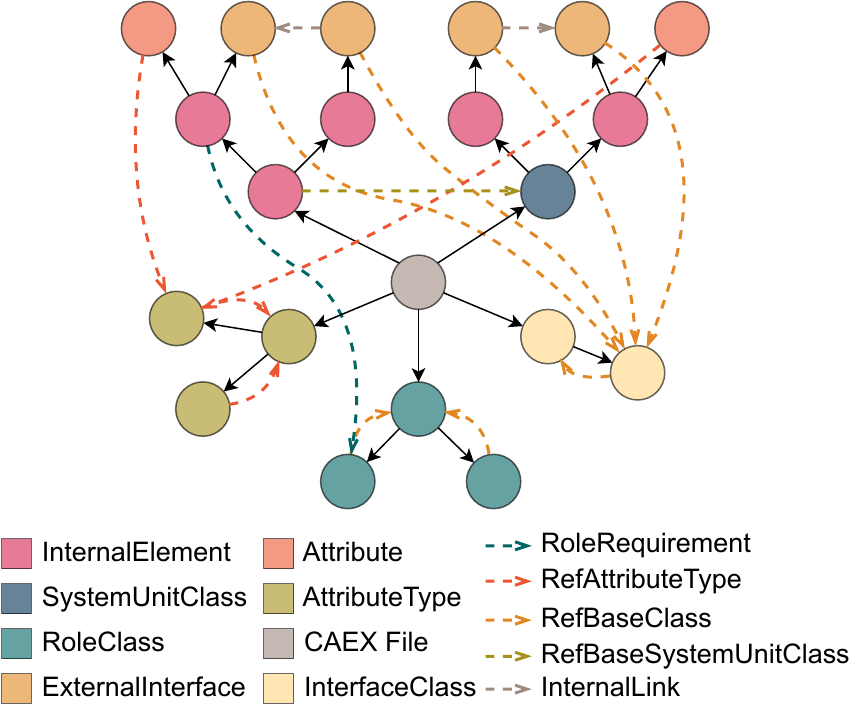}
\caption{AutomationML file from Figure \ref{fig:automationMLFile} visualized as a graph. Solid black lines are derived from the XML hierarchy, while dashed lines use ID- or path-based reference mechanisms. }
\label{fig:AmlRdfComparison}
\end{figure}

XML-based files can be considered as a specific subtype of graphs.
Here, the XML-Elements are considered as the nodes, whereas the hierarchical organization of elements forms the edges. 
If viewed in this way, XML files naturally take the shape of a Tree (s. solid black lines of Figure \ref{fig:AmlRdfComparison}). 
Trees are a specific type of acyclic, connected graph:
\begin{equation}\label{eq:Tree}
T = (V, E)
\end{equation}
where:
\begin{itemize}
    \item Each node has a unique path from the root, ensuring a clear hierarchical organization of data.
    \item Every edge is directed, so for each edge \( (u, v) \in E \), \( u \) is considered the \textit{parent} and \( v \) the \textit{child}.
\end{itemize}

XML's hierarchical nature ensure that the data can be traversed, queried, and manipulated as a tree, making it an ideal format for structured data representation. 
XML-based tools for querying and validation heavily rely on this hierarchical parent / child structure.

However, as stated in Section~\ref{sec:AutomationML}, AutomationML also allows additional references between nodes besides their hierarchical position in the XML file. 
Once the ReferencePaths and InternalLinks are considered, the AutomationML model no longer satisfies the property of having a unique path between every pair of nodes. An example of such a AutomationML graph can be seen in Figure~\ref{fig:AmlRdfComparison}.
It is therefore no longer a Tree as defined by Definition~\ref{eq:Tree}

The data instead forms a \textit{labeled directed multigraph}. It is defined as a triple 
\begin{equation}\label{eq:Multigraph}
G = (V, E, L),
\end{equation}
where:
\begin{itemize}
    \item[\( V \)] is a set of vertices (or nodes),
    \item[\( E \)] is a multiset of ordered pairs of \textit{directed edges}. Each edge is represented as \( (u, v) \), where \( u, v \in V \), with \( u \) as the \textit{source} and \( v \) as the \textit{target}, and
    \item[\( L \)] is a function that assigns a label to each edge in \( E \), mapping \( L : E \to \mathcal{L} \), where \( \mathcal{L} \) is a set of labels.
\end{itemize}
In this type of graph, multiple edges can exist between the same nodes. Edges are described further by their labels, that define which type of edge (e.g. RefBaseClass, InternalLink etc.) exists between the nodes. 
\medskip

For the use of XML tools on AutomationML files, this mismatch between XML's tree structure and AutomationML's multigraph structure results in some restrictions.

XML tools for queries (like xPath and xQuery) as well as for validation (like XSD and Schematron) were designed to handle hierarchical data in the shape of Trees (s. Equation~\ref{eq:Tree}). 
They therefore only offer limited support for connections that are not part of the XML tree structure (i.e. all dashed lines in Figure \ref{fig:AmlRdfComparison}). 

For data in the shape of labeled directed multigraphs, other data formats and technologies exist, that are able to natively deal with the data's cyclical and interconnected nature. 
Semantic Web Technologies are commonly used for this task and will be explained in more detail in the following Section. 

\subsection{Semantic Web Technologies}\label{sec:SWT}
Semantic Web Technologies are a framework of technologies designed to enable machines to meaningfully interpret and process data. The Resource Description Framework (RDF), its Schema Extension (RDFS) and the Web Ontology Language (OWL) are related technologies used in the context of the Semantic Web Technologies. They are often used together to represent and reason about knowledge in a machine-readable format.

RDF is a framework for representing information about resources on the web. It provides a standardized means to describe resources using subject-predicate-object triples. Each triple represents a statement or a fact about a resource.
Mathematically speaking, an RDF-Graph takes the shape of a labeled and directed multigraph, as it allows different types of directed edges.

The Web Ontology Language OWL, on the other hand, is a language for defining ontologies. An ontology is a formal description of concepts, relationships, and properties within a certain domain\cite{Jasper.1999}. 
While RDF provides a foundation for expressing relationships, OWL enhances the expressiveness and formalism of these relationships, enabling the creation of more advanced knowledge representations. 

The RDF Mapping Language (RML) is a mapping language defined to express customized mapping rules from heterogeneous data structures and serializations to the RDF data model\cite{Dimou.2014}. It has seen widespread use since its inception and can be applied to structured data sources like CSV, JSON or XML-based formats like AutomationML. 
Numerous platform independent and open source tools are available to automatically convert data sources to RDF-Triples, given a suitable mapping. 

SPARQL is a query language and protocol designed for querying and manipulating data stored in RDF format.
It can be used to express patterns and conditions in order to retrieve, construct, analyze, or modify information stored in RDF graphs\cite{w3c-sparql11}.

The Shapes Constraint Language (SHACL) is a language used for describing and validating the structure and content of data in RDF graphs. 
SHACL provides a powerful mechanism for defining and enforcing data validation rules in RDF graphs, helping ensure data quality and consistency in semantic web applications\cite{w3c-shacl}.

\section{Requirements and State of the Art}\label{sec:RequirementsAndSOTA}
In the past, numerous attempts have been made to create an AutomationML ontology. The following Section first outlines the requirements that a usable AutomationML ontology needs to fulfill. Afterwards it gives an overview of related works and previous AutomationML ontologies,  before comparing previous ontologies to the requirements. 

\subsection{Requirements}\label{sec:Requirements}
The resulting knowledge graph should be both highly expressive as well as easy to use for people that are already familiar with AutomationML's terminology. 
This results in a number of requirements, that influence later design decisions. 
In order to make the ontology easy to understand and use for AutomationML-practitioners, the terminology of the ontology should closely follow the established AutomationML terminology, resulting in Requirement~\ref{rq:AmlVocabulary}. 
Since queries as well as validation use cases make use of the entire width of information that is stored in an AutomationML file, the ontology should likewise cover the entire vocabulary that AutomationML has to offer (\ref{rq:AmlCompleteness}). 
However, as stated in Section \ref{sec:AutomationML}, the vocabulary within AutomationML is subject to contextual interpretation based on hierarchical positioning and inter-element relationships.
Recognizing these contextual semantics and making them explicitly available are important for both accurate modeling as well as usability of the resulting knowledge graph (\ref{rq:AmlSpecificCaex}).

AutomationML is an actively maintained and developed data exchange format. The ontology should therefore consider the most recent version of the AutomationML specification (\ref{rq:AmlVersion}). 

Most applications of Semantic Web Technologies make heavy use of OWL terminology concepts like class/property hierarchies and property characteristics. 
Besides following the AutomationML terminology, the ontology should therefore supplement the AutomationML concepts with the logically interpretable vocabulary used by OWL (\ref{rq:AmlSemanticLifting}). 

The terminological definitions in an AutomationML ontology usually consist of two parts: First, the 'core' AutomationML terminology, that defines the AutomationML metamodel (e.g. what a RoleClass is). This is needed to formally describe any AutomationML file. The second part is the individual classification hierarchy (e.g. which RoleClasses there are). These hierarchies change between projects, and can therefore not be modeled overarchingly for all AutomationML files. 
Since some translation effort for the creation and population of the ontology is therefore project specific, modeling 'just' the core AutomationML terminology in an ontology is not sufficient. 
Creating a full AutomationML ontology for a project thus necessitates a reusable mapping procedure (\ref{rq:AmlRdfMapping}).

\begin{table}[htbp]
\caption{Requirements for a functional AutomationML ontology.}
\label{tab:AML2OWLRequirements}
\setlength{\tabcolsep}{3pt}
\noindent\rule{\columnwidth}{0.4pt}
\begin{description}
    \item[R1.] Class/Property names should follow the AutomationML terminology.\label{rq:AmlVocabulary}
    \item[R2.] The ontology should cover the entire CAEX vocabulary.\label{rq:AmlCompleteness}
    \item[R3.] The ontology should allow explicit modeling of context-dependent AutomationML concepts.\label{rq:AmlSpecificCaex}
    \item[R4.] The ontology should follow the current AutomationML specification.\label{rq:AmlVersion}
    \item[R5.] Enrich and align concepts from AutomationML with OWL concepts.\label{rq:AmlSemanticLifting}
    \item[R6.] A mapping/translation approach for the automatic conversion from AutomationML files to the ontology must be available.\label{rq:AmlRdfMapping}
\end{description}

\noindent\rule{\columnwidth}{0.4pt}
\end{table}

\subsection{Prior AutomationML-Ontologies}\label{sec:priorAMLOntologies}
In the past, multiple attempts to create an AutomationML ontology have been published. 

Runde et al.\cite{Runde2009175} were the first to show the potential of combining CAEX with semantic web technologies. Their approach focuses mostly on representing CAEX using OWL concepts. However, the approach is lacking with regard to the representation of attributes, and deviates from the terminology that CAEX/AutomationML establish.  

Abele et al.\cite{L.Abele.2015} cover the most important aspects of the CAEX vocabulary, and even some basic context-dependent concepts like MirrorObjects. However, neither of those are complete as lots of additional information about the entities is not covered.

Grangel-Gonzales et al. \cite{GrangelGonzalez.2016} focus on the Attribute structure of AutomationML and are therefore lacking coverage of the remaining CAEX vocabulary. AutomationML files are mapped automatically using a XSLT-based tool. 

Kovalenko et al. \cite{O.Kovalenko.2015} focus on the description of the AutomationML vocabulary, but the authors do not completely cover the full AutomationML schema specification, particularly for data properties.

Kovalenko et al. later present an Ontology of the CAEX-submodel of the AutomationML-standard \cite{OlgaKovalenko.2018}. This version is essentially a combination of the previous works from \cite{GrangelGonzalez.2016} and \cite{O.Kovalenko.2015}. The ontology provides fairly good coverage of the basic CAEX concepts, with some notable exceptions like constraints. The context dependent concepts (like e.g. Mirror- or MappingObjects) remain missing.

Since the release of all considered ontologies, the AutomationML standard underwent continuous development and has since been released as Version 3.0 in 2022\cite{IEC62714-2:2022}. All ontologies were released before this date and can therefore not fulfill Requirement \ref{rq:AmlVersion}. Because of this, they lack concepts for the semantic AttributeLibraries introduced in \cite{IEC62714-2:2022}.

\begin{table}[htbp]
\caption{Requirement fulfillment of previous AutomationML Ontologies.}
\setlength{\tabcolsep}{3pt}
\begin{tabular}{|p{265pt}|c|c|c|c|c|}
\hline
Previous Ontology & \cite{L.Abele.2015} &  \cite{O.Kovalenko.2015} & \cite{GrangelGonzalez.2016} & \cite{OlgaKovalenko.2018} \\
\hline
R1: Terminology                         & \harveyball{50}   & \harveyball{50}   & \harveyball{50}   & \harveyball{50} \\
R2: Complete Coverage                 & \harveyball{50}       & \harveyball{50}   & \harveyball{0}   & \harveyball{100}     \\
R3: Context-dependent Entities        & \harveyball{50}       & \harveyball{0}       & \harveyball{0}       & \harveyball{0}     \\
R4: Current AutomationML Version           & \harveyball{0}       & \harveyball{0}       & \harveyball{0}       & \harveyball{0}     \\
R5: Enrich Model with OWL concepts & \harveyball{100}        & \harveyball{100}       & \harveyball{0}       & \harveyball{100}     \\
\hline
\end{tabular}
\label{tab:RequirementFulfillment}
\end{table}

As Table~\ref{tab:RequirementFulfillment} shows, none of the considered Ontologies fulfill all requirements. However, the ontology presented by Kovalenko et al. \cite{OlgaKovalenko.2018} achieves good coverage of the basic CAEX concepts. 
Since the new AutomationML version\cite{IEC62714-2:2022} introduces significant changes to the AutomationML vocabulary as well as the XML serialization, a new ontology needed to be developed. Since many of the basic CAEX concepts did not change, and \cite{OlgaKovalenko.2018} achieved good coverage in that area, their ontology was used as a starting point. 

Section~\ref{sec:AutomationMLontology} describes the content of this newly developed ontology.

\subsection{Prior Mapping Approaches}
Multiple publications approached the translation of AutomationML to OWL, using different target ontologies as well as translation technologies.

Out of the ontologies mentioned in Section \ref{sec:priorAMLOntologies}, Runde et al.\cite{Runde2009175}, Kovalenko et al.\cite{O.Kovalenko.2015} and Grangel-Gonzales et al.\cite{GrangelGonzalez.2016}mention an automatic translation. 
Runde et al.\cite{Runde2009175} propose a semi-automated mapping approach, but do not mention the implementation technology. Kovalenko et al. \cite{O.Kovalenko.2015} propose a transformation from AutomationML into OWL using the software framework \textit{Apache Jena}, which is unavailable. 
Finally, Grangel-Gonzales et al.\cite{GrangelGonzalez.2016} propose translating AutomationML to RDF using \textit{Krextor}, an XSLT-based framework for converting XML to RDF. However, \textit{Krextor} has not been maintained. 

Sabou et al. \cite{Sabou.2016} present the \textit{AutomationML-Analyzer}, which is a tool that showcases the functionalities made possible by the combination of AutomationML-files with semantic web technologies. 
The authors use the aforementioned AutomationML Ontology presented in \cite{O.Kovalenko.2015}. 
The mapping is performed using a two-step process, in which the AutomationML-files are first converted to Ecore, a domain model of the Eclipse Modeling Framework, and then converted from Ecore to OWL.
This conversion procedure is identified as less than ideal by the authors, as they propose a direct mapping as a potential future work.

With regard to mapping AutomationML files to OWL, Hua et al. \cite{Y.Hua.2019} introduce \textit{AutomationML concept models} for representing OWL complex classes in AutomationML. Additionally, the authors present algorithms for the bidirectional translation between OWL complex classes and their corresponding AutomationML concepts. Their focus, however, is on leveraging the semantics of the AutomationML concept models, not on providing practitioners with a straightforward method of translation between the two representations. 

\begin{table}[htbp]
\caption{Summary of previous approaches to translate AutomationML to OWL.}
\setlength{\tabcolsep}{3pt}
\begin{tabular}{|p{140pt}|c|c|c|c|}
\hline
Approach                                            & Technology    & Target Ontology               & Available     \\
\hline
Runde et al.\cite{Runde2009175}                     & -             & \cite{Runde2009175}           & \harveyball{0}       \\
Kovalenko et al.\cite{O.Kovalenko.2015}             & Apache Jena   & \cite{O.Kovalenko.2015}       & \harveyball{0}       \\
Grangel-Gonzales et al.\cite{GrangelGonzalez.2016}  & Krextor/XSLT  & \cite{GrangelGonzalez.2016}   & \harveyball{100}        \\
Sabou et al.\cite{Sabou.2016}                       & Ecore/Java    & \cite{O.Kovalenko.2015}       & \harveyball{100}       \\
Hua et al.\cite{Y.Hua.2019}                         & Custom/Java   & NA                            & \harveyball{100}       \\
\hline
\end{tabular}
\label{tab:mappingApproaches}
\end{table}

The preexisting mapping definitions target differ in their underlying technology as well as their target ontologies. As stated in Section \ref{sec:priorAMLOntologies}, these ontologies to not fully meet the outlined requirements, and will therefore be replaced by a new AutomationML ontology. 
Since the publication of these mapping approaches, \textit{RML} has emerged as the de facto standard declarative mapping language to automatically convert structured data into RDF triples, making it a natural choice for the implementation of a new mapping. 
Sections~\ref{sec:RmlMapping} and \ref{sec:UseCase} describe the mapping rules for the transformation as well as a detailed example. 
\section{AutomationML Ontology} \label{sec:AutomationMLontology}
To address the requirements outlined in Section~\ref{sec:Requirements}, a new ontology has been developed. This section introduces the AutomationML ontology by outlining its intended scope, presenting its main contents, and explaining the underlying design principles that guided its development.

\subsection{Ontology Scope}

AutomationML is structured around a clear separation between the metamodel and the model level.

The metamodel is defined by the CAEX schema (IEC 62424) together with AutomationML-specific constraints. It specifies the abstract language of AutomationML: which elements exist (InternalElement, SystemUnitClass, RoleClass, InterfaceClass, Attribute), how they may be related, and which properties they can carry. In ontological terms, this corresponds to the T-Box, i.e., the definition of the vocabulary and its structural rules. The core ontology introduced in this work captures precisely this metamodel.

On the model level, AutomationML distinguishes between two types of content:
\begin{itemize}
    \item Libraries (e.g., RoleClassLib, SystemUnitClassLib, InterfaceLib, AttributeTypeLib) define reusable class hierarchies. They extend the metamodel vocabulary with domain- or project-specific categories. In ontology terms, these libraries represent T-Box extensions, since they define new classes and relations, e.g., \textit{MixingTank} as a subclass of \textit{Tank}.
    \item Instances populate these libraries with concrete elements that describe a system. For example, a project may instantiate a particular \textit{MixingTank} as part of a specific plant. These correspond to A-Box assertions, since they describe individual objects and their relationships.
\end{itemize}

Because libraries and instances are typically user-defined and vary across projects, it is not feasible to provide a predefined ontology that covers them comprehensively. Instead, the ontology developed here focuses on the core vocabulary of AutomationML (metamodel, T-Box). Section~\ref{sec:RmlMapping} then specifies the rules for extending this core ontology with project-specific libraries (T-Box) and populating it with project instances (A-Box).

As described in Section~\ref{sec:priorAMLOntologies}, a new AutomationML ontology was designed that aims to to fulfill the Requirements outlined in Section~\ref{sec:Requirements}. The section first gives an overview of the CAEX vocabulary that could be reused (\ref{rq:AmlCompleteness}), before describing the concepts that had to be added to fulfill Requirements \ref{rq:AmlVersion} and \ref{rq:AmlSpecificCaex}.

\subsection{Ontology Summary}
As discussed in Section~\ref{sec:priorAMLOntologies}, the AutomationML ontology developed by Kovalenko et al.~\cite{OlgaKovalenko.2018} already provides substantial coverage of the core CAEX terms. This ontology was therefore chosen as the starting point for the development of the present AutomationML ontology.

Several modifications and extensions were applied to better align the ontology with current requirements:

\begin{itemize}
\item Terminology alignment: The terminology was adjusted to closely match the official AutomationML vocabulary (\cref{rq:AmlVocabulary}).
\item Completeness and specificity: The ontology was extended with previously unmodeled concepts from the CAEX metamodel (\cref{rq:AmlCompleteness}) and with context-dependent AutomationML concepts (\cref{rq:AmlSpecificCaex}).
\item Version updates: New concepts introduced in the latest AutomationML version, such as AttributeTypes, were incorporated (\cref{rq:AmlVersion}).
\item Documentation and annotations: Additional descriptions and annotations were added to improve clarity and usability.
\end{itemize}

\begin{figure*}[htbp]
\includegraphics[width=\textwidth]{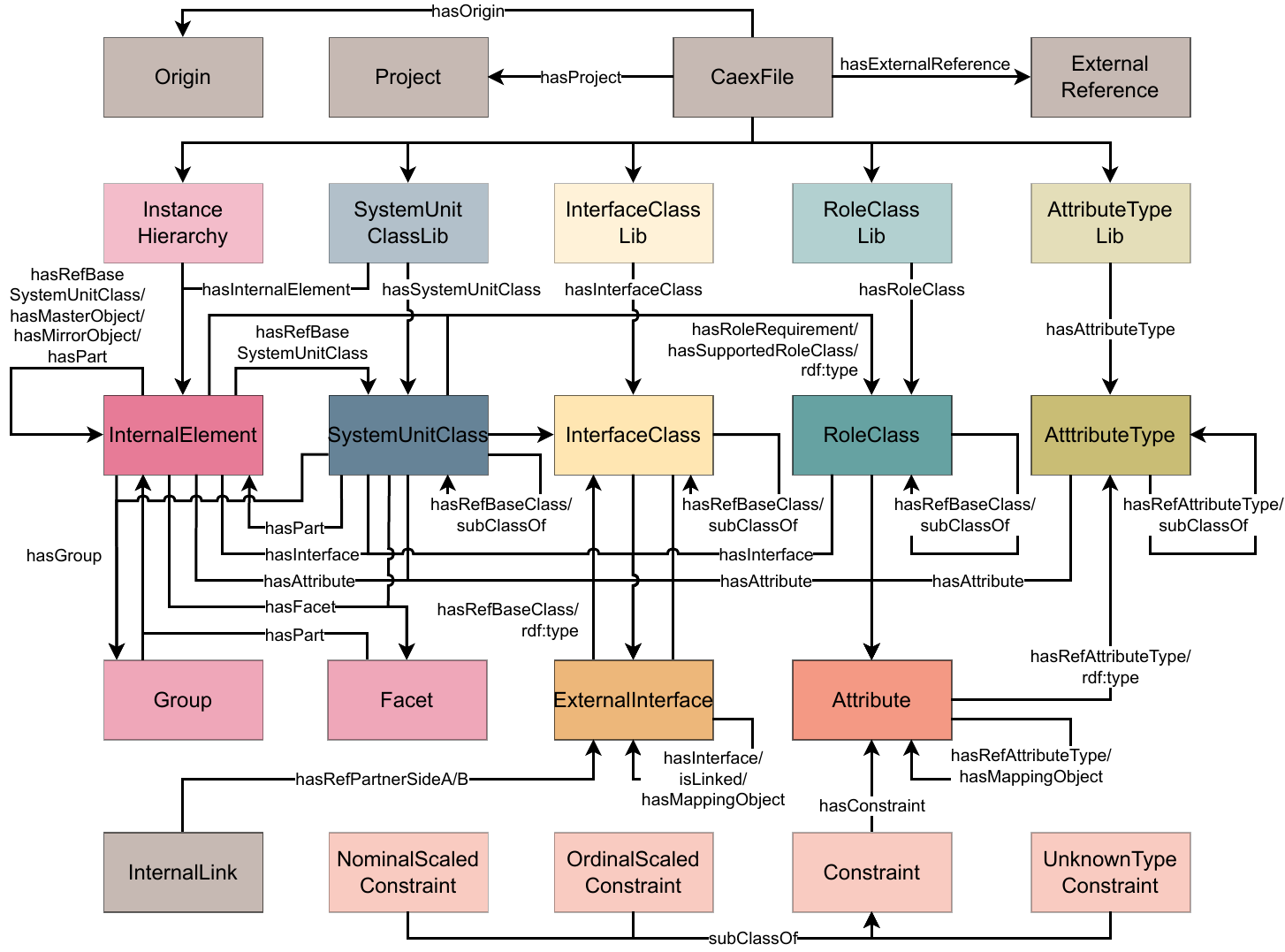}
\caption{Graphical overview of the classes and most common object properties of the AutomationML-Ontology.}
\label{fig:OntologyOverview}
\end{figure*}

The ontology follows the AutomationML metamodel and retains its original terminology. For instance, a RoleClass continues to reference its parent class using the \textit{hasRefBaseClass} object property. The corresponding ontology concept, \textit{rdfs:subClassOf}, is applied as a separate object property, in accordance with the mapping rules defined in Section~\ref{sec:RmlMapping}. 

A graphical overview of the classes and object properties in the AutomationML ontology is shown in Figure~\ref{fig:OntologyOverview}.

\subsection{Design Practices and Reuse}
The AutomationML ontology is published under W3ID, which is a permanent identifier, that is administered by the W3C Permanent Identifier Community Group. 
The ontology is permanently available using the ontology base IRI \textit{https://w3id.org/hsu-aut/AutomationML}. Administration and maintenance of the ontology is supported by the Institute of Automation Technology at Helmut-Schmidt-University.
Both the ontology itself as well as its classes and relationships are annotated with metadata documenting their creation process and correct usage. 
\section{Mapping AutomationML to OWL}\label{sec:RmlMapping}
While Section \ref{sec:AutomationMLontology} described the ontology that covers the AutomationML metamodel, the following section defines how the metamodel is extended by a class hierarchy.  With that aim, the following Section provides: 
\begin{enumerate}
    \item An alignment of AutomationML vocabulary with OWL-specific vocabulary (e.g. rdf:type and rdfs:subClassOf),
    \item A formal description of the relationships between CAEX, AutomationML, and OWL vocabulary,
    \item A formal description of context dependent AutomationML concepts,    
\end{enumerate}

The mapping rules in this section - as well as all other rules needed to automatically populate an AutomationML ontology with the data from an AutomationML file - are available in the RDF-Mapping-Language RML. 

\subsection{Creation of IRIs}\label{sec:CreationOfIRI}
Every resource inside an RDF graph has to be identified by an IRI, that distinguishes it from other resources. 
For AutomationML files, two main ways exist to create these identifiers: 
The first option relies on the IDs that are provided in the AutomationML file. While this approach is straightforward, IDs are only mandatory for InternalElements and ExternalInterfaces, where AutomationML also uses these IDs internally for references. 
For all other elements in AutomationML files, the use of IDs is optional and can therefore not reliably be used for referencing. 
Instead, elements without an ID are referenced through a reference path, which is the second option to create unique IRIs. 
In AutomationML, these elements must therefore have a name that is unique to their level in the element hierarchy.

In the RML mapping, this reference approach was mimicked: 
IDs are used as IRIs where they are mandatory (i.e. for InternalElements and ExternalInterfaces), while IRI-safe ReferencePaths are used for all remaining elements. 
References to AttributeTypes, Role-, SystemUnit- and InterfaceClasses take the form of reference paths as well, which makes the creation of ObjectProperties between instances and their respective classes straightforward. 
All instances are further given non-unique human-readable labels (via \emph{rdfs:label}) based on their names in AutomationML. 

\subsection{Creation of Classes and Instances}\label{sec:ClassesMapping}
For the mapping process, a decision had to be made which AutomationML concepts would be mapped as classes and as instances. 
For InternalElements and ExternalInterfaces, this decision is trivial: They are unique individuals, as evidenced by the mandatory use of an ID. 

Conceptually, classes in AutomationML are defined slightly different from classes in OWL. On one hand, they function like OWL classes - i.e. as a conceptual definition of a group of objects. 
On the other hand, classes in AutomationML can also function as a template for instances - i.e. they can posses attributes or be composed of multiple subcomponents. 
In this sense they show characteristics of both OWL classes as well as OWL instances. 

OWL 2 DL introduced the concept of \textit{punning}\cite{OWL2DL}. It allows repurposing the same identifier to represent both a class and an instance. 
Using this mechanism, AutomationML classes can be represented as OWL classes as well as OWL instances simultaneously. 
Punning is therefore used extensively in the mapping of RoleClasses, SystemUnitClasses, InterfaceClasses and AttributeTypes. 

In the class view, any RoleClass, SystemUnitClass, InterfaceClass, or AttributeType is converted to an OWL class. Class membership is implied, if instances reference the class using \textit{hasRefBaseSystemUnitClass}, \textit{hasRefBaseClass}, or \textit{hasRefAttributeType} respectively (s. Equation \ref{eq:Types}).

\begin{equation}\label{eq:Types}
\begin{split}
    \forall(A,B)\\
    (Attribute(A) \land \: hasRefAttributeType \:(A,B))\\
    \lor 
    (Interface(A) \land \: hasRefBaseClass \:(A,B))\\
    \lor 
    (InternalElement(A) \\
    \land \: hasRefBaseSystemUnitClass \:(A,B))\\    
    \implies \:rdf:type\:(A,B)
\end{split}
\end{equation}

In the instance view, any RoleClass, SystemUnitClass, InterfaceClass or AttributeType can be further described with any additional information from the AutomationML file. 
In this context, the OWL classes \textit{SystemUnitClass}, \textit{RoleClass}, \textit{InterfaceClass}, and \textit{AttributeType} are be treated as metaclasses, that define the structure and behavior of other classes. Classes defined in AutomationML are linked to their respective metaclasses using \textit{rdf:type}.

\subsection{Mapping of Roles}\label{sec:RolesMapping}
As described in Section \ref{sec:AutomationMLontology}, RoleClasses define the abstract functions of either InternalElements or SystemUnitClasses. 
AutomationML allows two types of RoleClass assignments: 
\begin{itemize}
    \item \textit{RoleRequirement} can be used to describe roles that an InternalElement or a SystemUnitClass is required to fill in a plant.
    \item \textit{SupportedRoleClass} can be used if an InternalElement or a SystemUnitClass is capable of filling a role. 
\end{itemize}
These relationships are modeled in two complementary mechanisms:
If an InternalElement or SystemUnitClass references a RoleClass via \textit{SupportedRoleClass}, it is considered to be an instance of that class (s. Equation~\ref{eq:TypeRole}). Additionally, the object properties \textit{hasRoleRequirement} and \textit{hasSupportedRoleClass} are assigned to explicitly distinguish between required and supported RoleClasses.

\begin{equation}\label{eq:TypeRole}
\begin{split}
    \forall(A,B)\\
    (InternalElement(A) \lor SystemUnitClass(A))\\ 
    \mspace{0mu}\land \:
    hasSupportedRoleClass \:(A,B)\\\implies \:rdf:type\:(A,B)
\end{split}
\end{equation}

\subsection{Mapping of Class Hierarchies}\label{sec:InheritanceMapping}
AutomationML allows modeling class hierarchies, in which subclasses can be derived from base classes. 
Inheritance between classes has to be explicitly modeled using \emph{RefAttributeType} for AttributeTypes or \emph{RefBaseClassPath} for SystemUnit-, Interface-, and RoleClasses. 

In the SystemUnitClass-, RoleClass-, InterfaceClass-, and AttributeTypeLibraries, this inheritance mechanism results in a class hierarchy using \emph{rdfs:sub-ClassOf} according to Equations~\ref{eq:ClassSUCICRC} and \ref{eq:ClassAT}. 

\begin{equation}\label{eq:ClassSUCICRC}
\begin{split}
    \forall(A,B)\\
    (SystemUnitClass(A) \lor InterfaceClass(A) \\
    \lor RoleClass(A))
    \land \:
    hasRefBaseClass \:(A,B)\\\implies \:rdfs:subClassOf\:(A,B)
\end{split}
\end{equation}

\begin{equation}\label{eq:ClassAT}
\begin{split}
    \forall(A,B)\\
    AttributeType(A) 
    \land \:
    hasRefAttributeType \:(A,B)\\
    \implies \:rdfs:subClassOf\:(A,B)
\end{split}
\end{equation}

\subsection{Mapping of Internal Structures}\label{sec:InternalStructures}
Some elements in AutomationML can  have an internal structure, i.e. they consist of multiple individual components. 
For example, both InternalElements and SystemUnitClasses can contain other InternalElements, Interfaces, and Attributes, that can, in turn, have an internal structure as well. 
In the same way, ExternalInterfaces can contain other ExternalInterfaces (e.g. a voltage interface on USB) and Attributes can contain other Attributes (e.g. \textit{x} and \textit{y} values on coordinates). 
In AutomationML, this relationship can be inferred from the XML-hierarchy in the source document. 
In the ontology, this implies a PartOf-Relationship. It is therefore modeled  using the object property \textit{hasPart}. 

\subsection{InternalLinks}\label{sec:InternalLinks}
In AutomationML, an InternalLink describes a directional connection between two interfaces. 
InternalLinks are crucial because they establish relationships between different components within a system model. 
The mapping uses this information in two ways: 
On one hand it follows the AutomationML terminology, in which an InternalLink instance has two RefPartnerSides. 
However, in comparison to AutomationML, OWL offers more suitable ways of representing links between elements. 
Additionally, an InternalLink is therefore represented using the :isLinked-Property that links both RefPartnerSides of an InternalLink while maintaining directionality. 

\begin{equation}\label{eq:InternalLink}
\begin{split}
    \forall(A,B,C)\\
    ExternalInterface(A) \land ExternalInterface(B) \\
    \mspace{0mu}\land \:  InternalLink(C) \\
    \mspace{0mu}\land \:  hasRefPartnerSideA(C, A) \\
    \mspace{0mu}\land \:  hasRefPartnerSideB(C, B)\\ 
    \implies isLinked\:(A,B) \\
\end{split}
\end{equation}

\subsection{Mirror-Objects}\label{sec:MirrorObjects}
Depending on the context, the RefBaseSystemUnitPath is used in AutomationML to describe two different relationships. If it connects an InternalElement with a SystemUnitClass, the InternalElement is based on the SystemUnitClass (s. Section \ref{sec:ClassesMapping}). 
However, if it connects an InternalElement with another InternalElement, it implies a Mirror-Relationship between the two elements. In this case, the second element follows all changes made to the first one. By introducing the :hasMirrorObject and :hasMasterObject, this connection is made explicit in the ontology. 
\begin{equation}\label{eq:MirrorObject}
\begin{split}
    \forall(A,B)\\
    InternalElement(A) \land InternalElement(B) \\
    \mspace{0mu}\land \:
    hasRefBaseSystemUnitClass(A, B)\\
    \implies (hasMasterObject\:(A,B) \\
    \land  \: hasMirrorObject \:(B,A))\\
\end{split}
\end{equation}

\subsection{Mapping-Objects}\label{sec:MappingObjects}
In AutomationML, MappingObjects are used to link and correlate elements from different models, ensuring interoperability and data consistency across various engineering domains and tools.
If e.g. an InternalElement and its RoleClass have corresponding Attributes or Interfaces with matching names, a MappingObject is implicitly assumed. 
If the names do not match, the relationship is explicitly stated in the  AutomationML file. In the latter case, the mapping is straightforward. 
In the former case, the mapping information has to be extracted from the data according to Equations~\ref{eq:MappingObject1}~/~\ref{eq:MappingObject2}.

\begin{equation}\label{eq:MappingObject1}
\begin{split}
    \forall(A,B,C,D,E,F)\\
    InternalElement(A) \land RoleClass(B) \\
    \land \:Attribute(C) \land Attribute(D) \\
    \land \:hasAttribute(A, C) \land hasAttribute(B, D)\\
    \land \:hasName(C, E) \land hasName(D, F) \\  
    \mspace{0mu}\land \:
    hasRoleRequirement(A, B)
    \mspace{0mu}\land \:
    Equal(E, F)\\
    \implies hasMappingObject\:(C,D) \\
\end{split}
\end{equation}

\begin{equation}\label{eq:MappingObject2}
\begin{split}
    \forall(A,B,C,D,E,F)\\
    InternalElement(A) \land RoleClass(B) \\
    \land \:Interface(C) \land Interface(D) \\
    \land \:hasInterface(A, C) \land hasInterface(B, D) \\
    \land \:hasName(C, E) \land hasName(D, F) \\  
    \mspace{0mu}\land \:
    hasRoleRequirement(A, B)
    \mspace{0mu}\land \:
    Equal(E, F)\\
    \implies hasMappingObject\:(C,D) \\
\end{split}
\end{equation}

\subsection{Facets and Groups}\label{sec:Facet}
Both Facets and Groups are not explicitly part of the original CAEX-Vocabulary. Instead, they are modeled as InternalElements that have a RoleRequirement of Group or Facet from the AutomationMLBaseRoleClassLibrary. Their membership in the class Group/Facet is therefore inferred according to Equations \ref{eq:Facet} and \ref{eq:Group}.

\begin{equation}\label{eq:Facet}
\begin{split}
    \forall(A,B)\\
    (InternalElement(A) \lor SystemUnitClass(A))\\
    \land InternalElement(B) 
    \land hasPart(A, B) \\    
    \land hasRoleRequirement(B, Facet) \\        
    \implies Facet(B) \:\land \:hasFacet\:(A,B)\\
\end{split}
\end{equation}

\begin{equation}\label{eq:Group}
\begin{split}
    \forall(A,B)\\
    (InternalElement(A) \lor SystemUnitClass(A))\\
    \land \:InternalElement(B) 
    \land \:hasPart(A, B) \\    
    \land \:hasRoleRequirement(B, Group) \\        
    \implies Group(B) \:\land :hasGroup\:(A,B)\\
\end{split}
\end{equation}

\subsection{Inheritance of Subcomponents}\label{sec:inheritanceSubcomponents}
In AutomationML, the internal structure of a class is inherited from a parent class to the child class. 
In the AutomationML model, this relationship is not explicitly modeled, but can instead be inferred through the inheritance relationship. 
It is additionally possible to override inherited objects.
For the mapping to RDF, this means that an internal class structure is inherited from the parent class if the inherited structure was not overridden (s. Equation~\ref{eq:Inheritance}).  

\begin{equation}\label{eq:Inheritance}
\begin{split}
    \forall(A,B,C,D,E)\\
    RoleClass(A) \land RoleClass(B) \\
    \mspace{0mu}\land \: hasRefBaseClass(A, B)\\
    \mspace{0mu}\land \: hasAttribute(B, C)
    \mspace{0mu}\land \: hasName(C, D)\\
    \mspace{0mu}\land \: \neg(\exists E \, (hasAttribute(A, E) \land hasName(E, D))) \\
    \implies hasAttribute\:(A,C) \\
\end{split}
\end{equation}

\subsection{Mapping Process and Usage}
Fully mapping an AutomationML file to RDF triples is a two-step process. In the first step, the provided RML mapping is executed. This populates the knowledge graph with all RDF triples corresponding to the contents of the file. In a second step, a set of SPARQL-Queries is executed, which adds additional triples, that cannot otherwise be created by means of pure RML mapping. 
This information includes the inheritance of Attributes and Interfaces from a parent element, as well as creating explicit MappingObjects of Attributes and Interfaces. 

A mapping application is available\footnote{https://github.com/hsu-aut/aml2owl}, that allows automatic mapping of a file from the command line. It includes both the RML mapping as well as the SPARQL queries. 

Alternatively, the mapping and the SPARQL queries are individually available on GitHub\footnote{https://github.com/tomwestermann/AutomationML2Ontology}. The mapping can be executed using any RML-compliant mapping tool, while the SPARQL queries can be executed using any compliant SPARQL processing tool. 
\section{Mapping Example}\label{sec:UseCase}
To illustrate the mapping rules from the previous section, a Four-Tank Mixing System will be used as a running example. 
Further models and test cases are available on Github.

\subsection{System Description: Four-Tank Mixing System}\label{sec:ExampleSystem}
\begin{figure}[htbp]
\includegraphics[width=\linewidth]{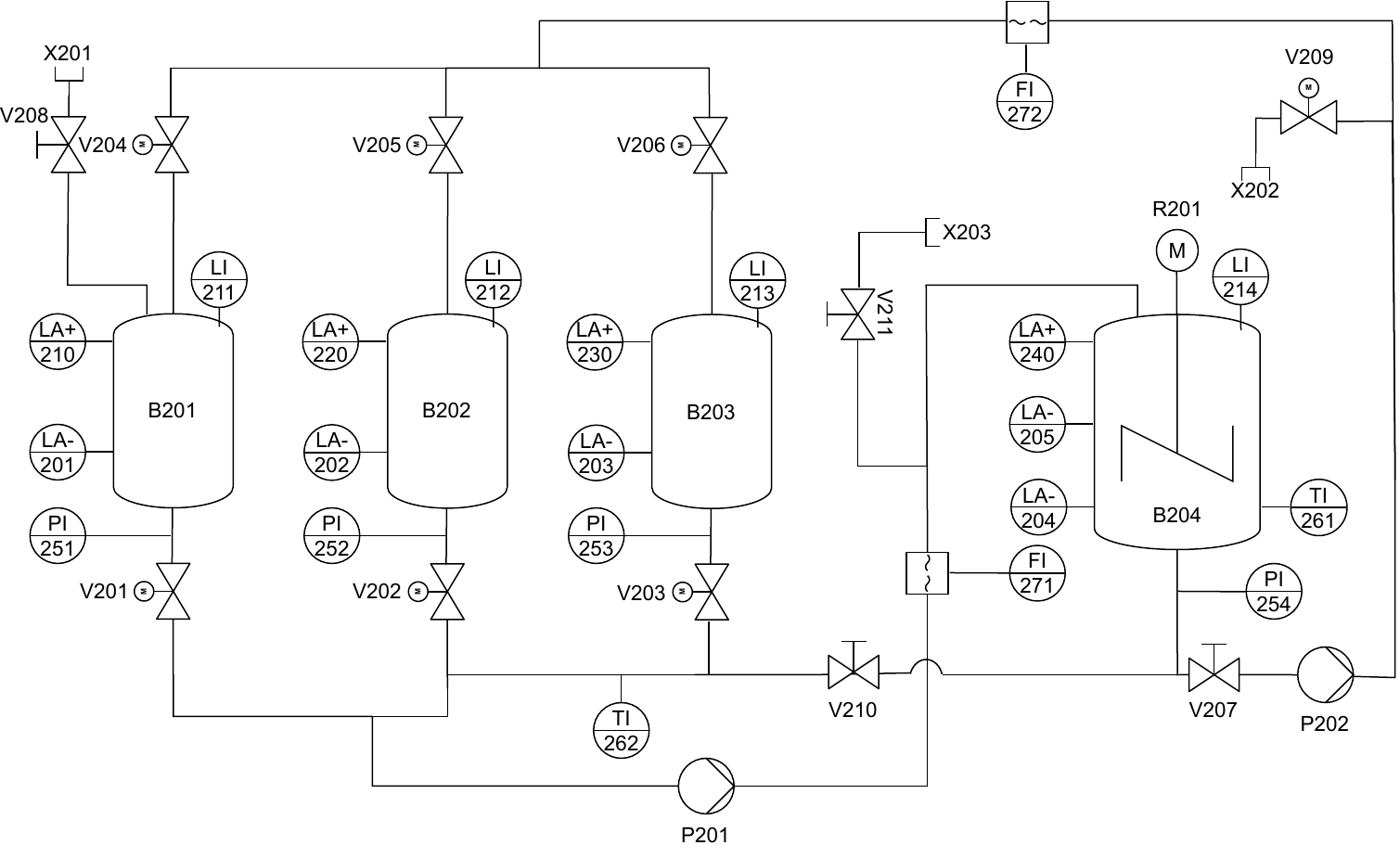}
\caption{P\&ID of the Fluid Mixing System.}
\label{fig:MixingPIDFlowDiagram}
\end{figure}

In the main use case, a laboratory scale process plant in the Institute of Automation Technology at Helmut-Schmidt-University is considered. It is used for fluid mixing and consists of three dosing tanks, a mixing tank, eleven valves and two pumps. Pipes connect the various components for fluid transport. 
A Piping and Instrumentation Diagram (P\&ID) of the system can be seen in Figure \ref{fig:MixingPIDFlowDiagram}. The system's topology was described in AutomationML according to the best practice recommendations by the AutomationML society.

The remainder of this section shows how the main concepts of AutomationML are translated to RDF-Triples using excerpts from the AutomationML file of this system.

\subsection{Creation of IRIs}
As described in Section \ref{sec:CreationOfIRI}, the creation of IRIs for the ontology follows the reference mechanisms that AutomationML is using internally. 
Therefore, IDs are only used for the IRI creation of InternalElements and ExternalInterface.
For all other elements, unique paths are created using the "Name"-Attribute. This mimics AutomationMLs approach to references.

\begin{figure}[ht]
\includegraphics[width=\linewidth]{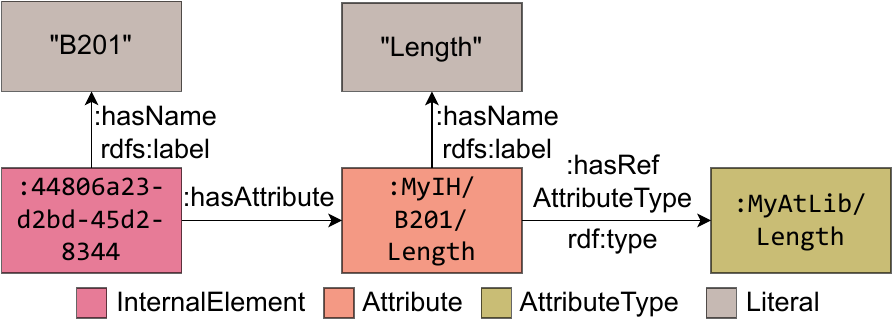}
\caption{RDF representation of the AutomationML content from Listing \ref{list:IRIexample}.}
\label{fig:IRIexample}
\end{figure}

Listing~\ref{list:IRIexample} and Figure~\ref{fig:IRIexample} show an example of this logic. 
While the InternalElement \textit{B201} is assigned an IRI from its ID, both the Attribute \textit{Length} as well as the AttributeType \textit{Length} use their paths as an IRI. 
\begin{listing}[ht]
\caption{\sffamily\fontencoding{T1}\fontseries{m}\fontsize{7}{8.4}\selectfont Example of an InternalElement with an Attribute "Length".}
\label{list:IRIexample}
\begin{Verbatim}[commandchars=\\\{\}]
\PYG{n+nt}{\PYGZlt{}InstanceHierarchy}\PYG{+w}{ }\PYG{n+na}{Name=}\PYG{l+s}{\PYGZdq{}MyIH\PYGZdq{}}\PYG{n+nt}{\PYGZgt{}}
\PYG{+w}{ }\PYG{n+nt}{\PYGZlt{}InternalElement}\PYG{+w}{ }\PYG{n+na}{Name=}\PYG{l+s}{\PYGZdq{}B201\PYGZdq{}}
\PYG{+w}{  }\PYG{n+na}{ID=}\PYG{l+s}{\PYGZdq{}44806a23\PYGZhy{}d2bd\PYGZhy{}45d2\PYGZhy{}8344\PYGZdq{}}\PYG{n+nt}{\PYGZgt{}}
\PYG{+w}{  }\PYG{n+nt}{\PYGZlt{}Attribute}\PYG{+w}{ }\PYG{n+na}{Name=}\PYG{l+s}{\PYGZdq{}Length\PYGZdq{}}
\PYG{+w}{   }\PYG{n+na}{RefAttributeType=}\PYG{l+s}{\PYGZdq{}MyAtLib/Length\PYGZdq{}}\PYG{n+nt}{\PYGZgt{}}
\PYG{+w}{  }\PYG{n+nt}{\PYGZlt{}/Attribute\PYGZgt{}}
\PYG{+w}{ }\PYG{n+nt}{\PYGZlt{}/InternalElement\PYGZgt{}}
\PYG{n+nt}{\PYGZlt{}/InstanceHierarchy\PYGZgt{}}
\end{Verbatim}
\end{listing}

\subsection{Creation of Classes and Instances}
The following section provides an example of the creation of classes and instances. 
Listing~\ref{list:ClassesInstances} shows an example of an InternalElement, that has all major elements that AutomationML uses: A SystemUnitClass, an Attribute, an ExternalInterface as well as two RoleClasses - one as a SupportedRoleClass and one as a RoleRequirement. 

\begin{listing}[ht]
\caption{\sffamily\fontencoding{T1}\fontseries{m}\fontsize{7}{8.4}\selectfont InternalElement with all basic AutomationML elements.
}
\label{list:ClassesInstances}
\begin{Verbatim}[commandchars=\\\{\}]
\PYG{n+nt}{\PYGZlt{}InternalElement}\PYG{+w}{ }\PYG{n+na}{Name=}\PYG{l+s}{\PYGZdq{}B201\PYGZdq{}}
\PYG{+w}{ }\PYG{n+na}{ID=}\PYG{l+s}{\PYGZdq{}44806a23\PYGZhy{}d2bd\PYGZhy{}45d2\PYGZhy{}8344\PYGZdq{}}
\PYG{+w}{ }\PYG{n+na}{RefBaseSystemUnitPath=}\PYG{l+s}{\PYGZdq{}MySucLib/Vessel\PYGZdq{}}\PYG{n+nt}{\PYGZgt{}}
\PYG{+w}{ }\PYG{n+nt}{\PYGZlt{}Attribute}\PYG{+w}{ }\PYG{n+na}{Name=}\PYG{l+s}{\PYGZdq{}Length\PYGZdq{}}
\PYG{+w}{  }\PYG{n+na}{RefAttributeType=}\PYG{l+s}{\PYGZdq{}MyAtLib/Length\PYGZdq{}}\PYG{n+nt}{\PYGZgt{}}
\PYG{+w}{ }\PYG{n+nt}{\PYGZlt{}/Attribute\PYGZgt{}}
\PYG{+w}{ }\PYG{n+nt}{\PYGZlt{}ExternalInterface}\PYG{+w}{ }\PYG{n+na}{Name=}\PYG{l+s}{\PYGZdq{}Input\PYGZdq{}}
\PYG{+w}{  }\PYG{n+na}{ID=}\PYG{l+s}{\PYGZdq{}ce45d7b3\PYGZhy{}169d\PYGZhy{}4be8\PYGZhy{}9746\PYGZdq{}}
\PYG{+w}{  }\PYG{n+na}{RefBaseClassPath=}\PYG{l+s}{\PYGZdq{}MyIcLib/Port\PYGZdq{}}\PYG{n+nt}{\PYGZgt{}}
\PYG{+w}{ }\PYG{n+nt}{\PYGZlt{}/ExternalInterface\PYGZgt{}}
\PYG{+w}{ }\PYG{n+nt}{\PYGZlt{}SupportedRoleClass}
\PYG{+w}{  }\PYG{n+na}{RefRoleClassPath=}\PYG{l+s}{\PYGZdq{}MyRcLib/Vessel\PYGZdq{}}\PYG{+w}{ }\PYG{n+nt}{/\PYGZgt{}}
\PYG{+w}{ }\PYG{n+nt}{\PYGZlt{}RoleRequirements}
\PYG{+w}{  }\PYG{n+na}{RefBaseRoleClassPath=}\PYG{l+s}{\PYGZdq{}MyRcLib/DosingTank\PYGZdq{}}\PYG{+w}{ }\PYG{n+nt}{/\PYGZgt{}}
\PYG{+w}{ }\PYG{n+nt}{\PYGZlt{}/RoleRequirements\PYGZgt{}}
\PYG{n+nt}{\PYGZlt{}/InternalElement\PYGZgt{}}
\end{Verbatim}
\end{listing}

\begin{figure}[ht]
\includegraphics[width=\linewidth]{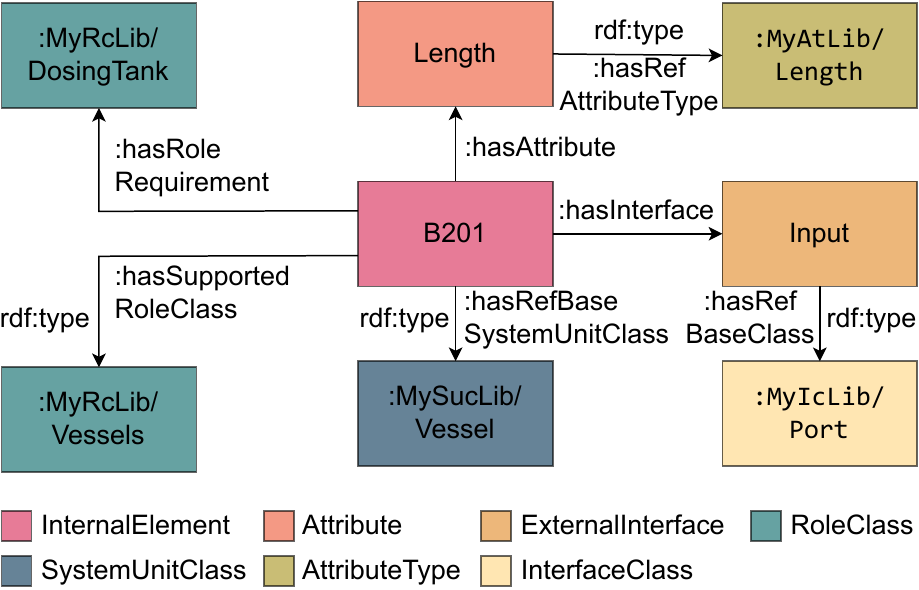}
\caption{Excerpt of resulting RDF-Triples for Classes and Instances from Listing \ref{list:ClassesInstances}.}
\label{fig:AML2OWLInstancesClasses}
\end{figure}

Figure \ref{fig:AML2OWLInstancesClasses} shows the corresponding relationship between OWL instances and classes from Listing \ref{list:ClassesInstances}: 
For example, the InternalElement \textit{B201} is:
\begin{itemize}
    \item an instance of the SystemUnitClass \textit{Vessel}, since it is its RefBaseSystemUnitClass,
    \item an instance of the RoleClass \textit{Vessels}, since it supports that RoleClass,
    \item not an instance of the RoleClass \textit{DosingTank}, since having a RoleRequirement does not imply class membership. 
\end{itemize}
The Attribute \textit{Length} and the Interface \textit{Input} are also instances of their respective AttributeTypes and BaseClasses. 
The RDF representation of the AutomationML data additionally connects the instances to their classes using the AutomationML terminology. 

\subsection{Mapping of Class Hierarchies}\label{MappingClassHierarchies}
The previous section showed how an InternalElement (or its Attributes and Interfaces) can relate to OWL classes that are derived from AutomationML libraries. 

Since AutomationML allows inheritance, these OWL classes become part of a class hierarchy. 
AutomationML additionally provides further meta information about these classes (e.g. default values, semantic references, provenance, etc.).
To adequately capture this information,  punning, i.e. the simultaneous use of the same IRI as a class and an instance, is used. 

\begin{figure}[ht]
\includegraphics[width=\linewidth]{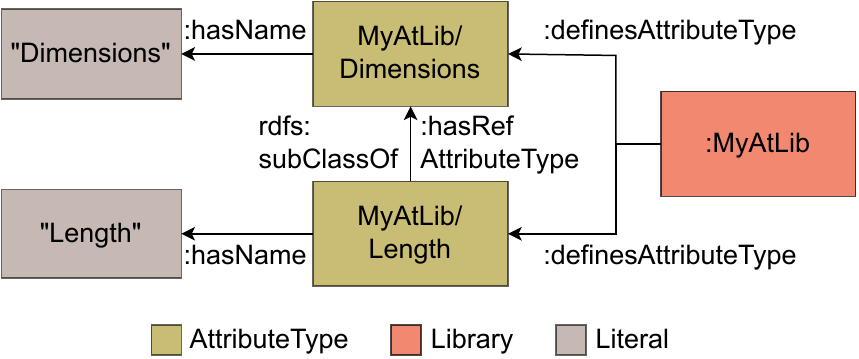}
\caption{Excerpt of resulting RDF-Triples from Listing \ref{list:ClassHierarchies}.}
\label{fig:ClassHierarchies}
\end{figure}

The example in Listing~\ref{list:ClassHierarchies} and Figure \ref{fig:ClassHierarchies} shows how the AttributeType \textit{Length} from the previous example is mapped and embedded into an OWL class hierarchy. For AttributeTypes, the \textit{rdfs:subClassOf}-relationship can be inferred from the \textit{RefAttributeType} in AutomationML.

\begin{listing}[ht]
\caption{\sffamily\fontencoding{T1}\fontseries{m}\fontsize{7}{8.4}\selectfont \textcolor{black}{Excerpt from the AttributeLibrary.}}
\label{list:ClassHierarchies}
\begin{Verbatim}[commandchars=\\\{\}]
\PYG{n+nt}{\PYGZlt{}AttributeTypeLib}\PYG{+w}{ }\PYG{n+na}{Name=}\PYG{l+s}{\PYGZdq{}MyAtLib\PYGZdq{}}\PYG{n+nt}{\PYGZgt{}}
\PYG{+w}{ }\PYG{n+nt}{\PYGZlt{}AttributeType}
\PYG{+w}{  }\PYG{n+na}{Name=}\PYG{l+s}{\PYGZdq{}Dimensions\PYGZdq{}}\PYG{n+nt}{\PYGZgt{}}
\PYG{+w}{ }\PYG{n+nt}{\PYGZlt{}AttributeType}\PYG{+w}{ }\PYG{n+na}{Name=}\PYG{l+s}{\PYGZdq{}Length\PYGZdq{}}
\PYG{+w}{  }\PYG{n+na}{RefAttributeType=}\PYG{l+s}{\PYGZdq{}MyAtLib/Dimensions\PYGZdq{}}\PYG{+w}{ }\PYG{n+nt}{/\PYGZgt{}}
\PYG{+w}{ }\PYG{n+nt}{\PYGZlt{}/AttributeType\PYGZgt{}}
\PYG{n+nt}{\PYGZlt{}/AttributeTypeLib\PYGZgt{}}
\end{Verbatim}
\end{listing}   

OWL Class hierarchies are created from all AutomationML libraries, i.e. also from the RoleClass-, SystemUnitClass-, and InterfaceClassLibraries. The mapping approach is similar for all other libraries.

\subsection{Mapping of InternalLinks}
\begin{figure}[ht]
\includegraphics[width=\linewidth]{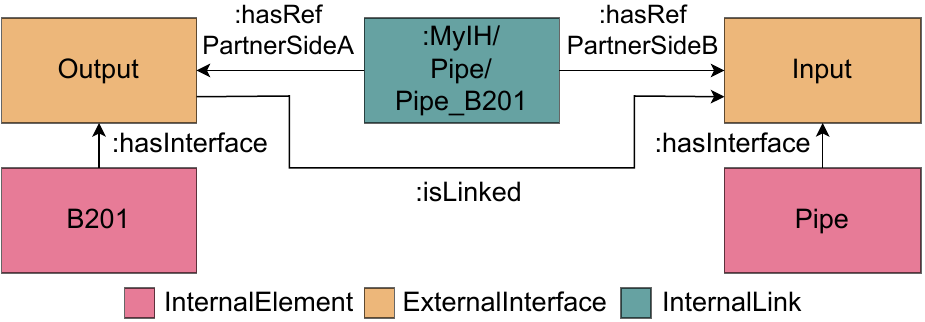}
\caption{Resulting RDF-Graph from Listing \ref{list:InternalLink} of two InternalElements with two linked Interfaces.}
\label{fig:InternalLink}
\end{figure}

Listing \ref{list:InternalLink} shows an example of two InternalElements, whose interfaces are connected by an InternalLink. 
Figure \ref{fig:InternalLink} shows both mechanisms of linking their Interfaces. On one hand, they are connected in a way that very closely follows the AutomationML-vocabulary with an instance of the class InternalLink that connects the Interfaces through the Object Property :hasRefPartnerSideA/B. On the other hand, they are connected directly through the object property :isLinked. This makes traversing the connections between InternalElements (e.g. for queries) more intuitive, while preserving the terminology most familiar to AutomationML practitioners. 

\begin{listing}[ht]
\caption{\sffamily\fontencoding{T1}\fontseries{m}\fontsize{7}{8.4}\selectfont Example of an InternalLink that connects a Pipe with a Tank.}
\label{list:InternalLink}
\begin{Verbatim}[commandchars=\\\{\}]
\PYG{n+nt}{\PYGZlt{}InstanceHierarchy}\PYG{+w}{ }\PYG{n+na}{Name=}\PYG{l+s}{\PYGZdq{}MyIH\PYGZdq{}}\PYG{n+nt}{\PYGZgt{}}
\PYG{+w}{ }\PYG{n+nt}{\PYGZlt{}InternalElement}\PYG{+w}{ }\PYG{n+na}{Name=}\PYG{l+s}{\PYGZdq{}Pipe\PYGZdq{}}
\PYG{+w}{  }\PYG{n+na}{ID=}\PYG{l+s}{\PYGZdq{}a20e3eac\PYGZhy{}9ec0\PYGZhy{}41f1\PYGZhy{}852a\PYGZdq{}}\PYG{n+nt}{\PYGZgt{}}
\PYG{+w}{  }\PYG{n+nt}{\PYGZlt{}ExternalInterface}\PYG{+w}{ }\PYG{n+na}{Name=}\PYG{l+s}{\PYGZdq{}Output\PYGZdq{}}
\PYG{+w}{   }\PYG{n+na}{ID=}\PYG{l+s}{\PYGZdq{}6eea7a40\PYGZhy{}43fd\PYGZhy{}4aee\PYGZhy{}a1d3\PYGZhy{}\PYGZdq{}}\PYG{n+nt}{\PYGZgt{}}
\PYG{+w}{  }\PYG{n+nt}{\PYGZlt{}/ExternalInterface\PYGZgt{}}
\PYG{+w}{  }\PYG{n+nt}{\PYGZlt{}InternalLink}\PYG{+w}{ }\PYG{n+na}{Name=}\PYG{l+s}{\PYGZdq{}Pipe\PYGZus{}B201\PYGZdq{}}
\PYG{+w}{   }\PYG{n+na}{RefPartnerSideA=}\PYG{l+s}{\PYGZdq{}6eea7a40\PYGZhy{}43fd\PYGZhy{}4aee\PYGZhy{}a1d3\PYGZdq{}}
\PYG{+w}{   }\PYG{n+na}{RefPartnerSideB=}\PYG{l+s}{\PYGZdq{}ce45d7b3\PYGZhy{}169d\PYGZhy{}4be8\PYGZhy{}9746\PYGZdq{}}\PYG{n+nt}{/\PYGZgt{}}
\PYG{+w}{ }\PYG{n+nt}{\PYGZlt{}/InternalElement\PYGZgt{}}
\PYG{+w}{ }\PYG{n+nt}{\PYGZlt{}InternalElement}\PYG{+w}{ }\PYG{n+na}{Name=}\PYG{l+s}{\PYGZdq{}B201\PYGZdq{}}
\PYG{+w}{  }\PYG{n+na}{ID=}\PYG{l+s}{\PYGZdq{}44806a23\PYGZhy{}d2bd\PYGZhy{}45d2\PYGZhy{}8344\PYGZdq{}}\PYG{n+nt}{\PYGZgt{}}
\PYG{+w}{  }\PYG{n+nt}{\PYGZlt{}ExternalInterface}\PYG{+w}{ }\PYG{n+na}{Name=}\PYG{l+s}{\PYGZdq{}Input\PYGZdq{}}
\PYG{+w}{   }\PYG{n+na}{ID=}\PYG{l+s}{\PYGZdq{}ce45d7b3\PYGZhy{}169d\PYGZhy{}4be8\PYGZhy{}9746\PYGZdq{}}\PYG{n+nt}{\PYGZgt{}}
\PYG{+w}{  }\PYG{n+nt}{\PYGZlt{}/ExternalInterface\PYGZgt{}}
\PYG{n+nt}{\PYGZlt{}/InstanceHierarchy\PYGZgt{}}
\end{Verbatim}
\end{listing}

\section{Application}\label{sec:Usage}
A reliable and reproducible way of translating AutomationML-Files to RDF-Triples opens up a plethora of new applications using the different technologies from the semantic web stack (s. Section \ref{sec:SWT}).
In this section, two such applications, namely querying and validation, and their advantages will be described. 
The first application describes the use of the query language SPARQL to construct complex graph queries. The capabilities of SPARQL are compared to xPath and xQuery, which are considered the most common query languages that operate directly on the XML representation of AutomationML. 
The second application describes an automatic validation of an AutomationML file by using SHACL to formulate validation patterns. SHACL's capabilities are then compared to XSD and Schematron, two common approaches to validate the content of XML files. 

\subsection{Complex Graph Queries}\label{sec:complexGraphQueries}
As stated in Section~\ref{sec:GraphTheory}, the AutomationML model describes a labeled and directed multigraph, that is serialized as an XML file in the shape of a Tree. 

Common tools to perform queries on XML data (like xPath or xQuery) were designed to navigate tree structures, and are therefore lacking mechanisms to handle the additional edges in the graph created by ReferencePaths and InternalLinks. 

However, many industrial applications could significantly benefit from the possibility to query the full spectrum of information provided by the Automation-ML-file. 

To illustrate the advantages of such graph queries, we revisit the Four-Tank-Mixing System described in Section \ref{sec:ExampleSystem}. 
We will compare the extraction of information using different approaches: xPath and xQuery, which operate directly on the XML logic underlying the AutomationML file, and SPARQL, which queries an RDF graph generated from the AutomationML file.

Sets of search queries of varying complexity will be considered to highlight the strengths and specific limitations of each approach, demonstrating the potential benefits of employing graph queries in relevant scenarios.

\subsubsection*{Use Case 1: Conditional Element-Selection}
\begin{figure}[ht]
\includegraphics[width=\linewidth]{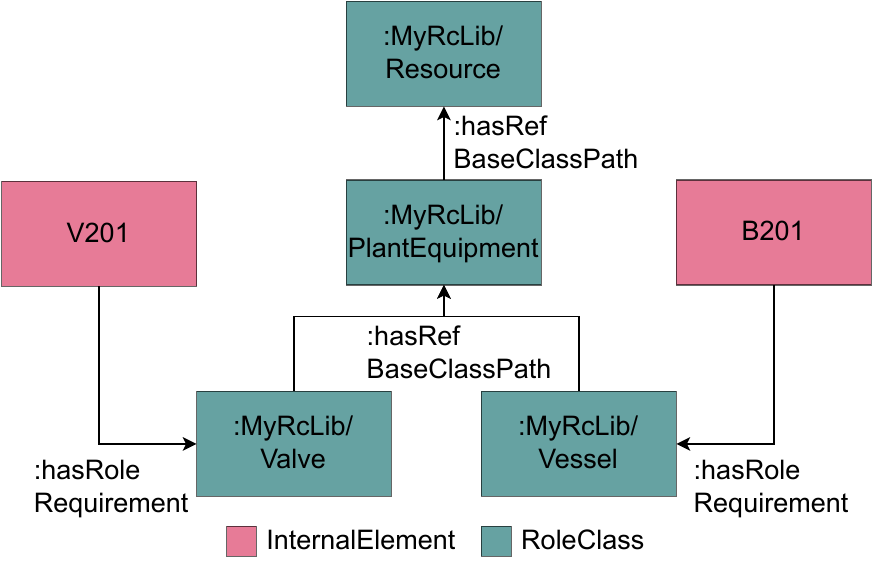}
\caption{Two InternalElements with a RoleClass Hierarchy.}
\label{fig:SimpleQueryExample}
\end{figure}

In the first Use Case, the simple selection of elements based on some selection criteria was considered. 
Three different scenarios were designed. 
Each scenario differs in the degree of separation in the transformed graph between the element $A$ that should be selected and the element $X$ at which the selection criteria is applied. 
\begin{enumerate}
    \item \textit{Scenario 1 - First Degree Selection}: 
    
    Select all InternalElements A that have a certain RoleRequirement B:    
    \begin{flushleft}
        $\:\:\:\:InternalElement(A) \linebreak \land hasRoleRequirement(A,B)$
    \end{flushleft}
    \item \textit{Scenario 2 - Second Degree Selection}: 
    
    Select all InternalElements A that have a RoleRequirement B, where B references a RoleClass C as a RefBaseClass:
    \begin{flushleft}   
        $\:\:\:\:InternalElement(A) \linebreak \land hasRoleRequirement(A,B) \linebreak \land hasRefBaseClass(B,C)$
    \end{flushleft}
    \item \textit{Scenario 3 - nth Degree Selection}: 
    
    Select all InternalElements A that have a RoleRequirement B, where B references a RoleClass C as a RefBaseClass through transitive closure. 
    \begin{flushleft}
        $\:\:\:\:InternalElement(A) \linebreak \land hasRoleRequirement(A,B) \linebreak \land hasRefBaseClass*(B,C)$
    \end{flushleft}
\end{enumerate}
Example queries of these three scenarios were applied to the  the Mixing System: All InternalElements should be selected that have the RoleRequirement of being 1) a Vessel, 2) a Plant Equipment (or any direct descendant) or 3) a Resource (or any transitive descendant). For the example in Figure \ref{fig:SimpleQueryExample}, the first scenario should return just \textit{B201}, while the queries from the other scenarios should return both InternalElements. 

For all three of these Scenarios, queries were written in each of the three query languages. 

\begin{listing}[ht]
\caption{\sffamily\fontencoding{T1}\fontseries{m}\fontsize{7}{8.4}\selectfont xPath, xQuery and SPARQL Query for Scenario 1.}
\label{list:QueriesScenario1}
\center{\textbf{xPath:}}
\begin{Verbatim}[commandchars=\\\{\}]
\PYG{p}{//}\PYG{n+nt}{InternalElement}\PYG{p}{[}\PYG{n+nt}{RoleRequirements}\PYG{p}{[}
\PYG{+w}{  }\PYG{n+na}{@RefBaseRoleClassPath}\PYG{o}{=}\PYG{l+s+s1}{\PYGZsq{}MyRcLib/Vessel\PYGZsq{}}\PYG{p}{]]}
\end{Verbatim}
\textbf{xQuery:}
\begin{Verbatim}[commandchars=\\\{\}]
\PYG{k}{for}\PYG{+w}{ }\PYG{n+nv}{\PYGZdl{}}\PYG{n}{ie}\PYG{+w}{ }\PYG{o+ow}{in}\PYG{+w}{ }\PYG{p}{//}\PYG{n+nt}{InternalElement}
\PYG{+w}{  }\PYG{k}{where}\PYG{+w}{ }\PYG{n+nv}{\PYGZdl{}}\PYG{n}{ie}\PYG{p}{/}\PYG{n+nt}{RoleRequirements}\PYG{p}{/}
\PYG{+w}{    }\PYG{n+na}{@RefBaseRoleClassPath}\PYG{+w}{ }\PYG{o}{=}\PYG{+w}{ }\PYG{l+s+s2}{\PYGZdq{}MyRcLib/Vessel\PYGZdq{}}
\PYG{+w}{  }\PYG{k}{return}\PYG{+w}{ }\PYG{n+nv}{\PYGZdl{}}\PYG{n}{ie}
\end{Verbatim}
\textbf{SPARQL:}
\begin{Verbatim}[commandchars=\\\{\}]
\PYG{k}{SELECT} \PYG{n+nv}{?ie}
\PYG{k}{WHERE} \PYG{p}{\PYGZob{}}
  \PYG{n+nv}{?ie} \PYG{n+nn}{rdf}\PYG{p}{:}\PYG{n+nt}{type} \PYG{n+nn}{aml}\PYG{p}{:}\PYG{n+nt}{InternalElement} \PYG{p}{.}
  \PYG{n+nv}{?ie} \PYG{n+nn}{aml}\PYG{p}{:}\PYG{n+nt}{hasRoleRequirement} \PYG{n+nn}{aml}\PYG{p}{:}\PYG{n+nt}{MyRcLib}\PYG{o}{/}\PYG{n+nt}{Vessel}\PYG{p}{.\PYGZcb{}}
\end{Verbatim}
\end{listing}

The queries regarding the first scenario were easily achievable in all three languages (s. Listing~\ref{list:QueriesScenario1}).

However, both the second and third scenario make use of the AutomationML reference paths. 
The query in Scenario 2 not only references the RoleRequirement of InternalElement A, but also the RoleRequirement's RefBaseClass. 
In the XML tree, this information is located at a different branch of the tree at the RoleRequirement node and thus necessitates a traversal of the graph that does not follow the hierarchical structure of the XML file. 
Scenario 3 further extends this by combining the filter condition with an open-ended transitive closure statement, i.e. a sequence of RefBaseClassPaths of indeterminate length. 

\begin{listing}[ht]
\caption{\sffamily\fontencoding{T1}\fontseries{m}\fontsize{7}{8.4}\selectfont xQuery for Scenario 2 and SPARQL Query for Scenario 3.}
\label{list:QueriesScenario23}
\center{\textbf{xQuery:}}
\begin{Verbatim}[commandchars=\\\{\}]
\PYG{k}{let}\PYG{+w}{ }\PYG{n+nv}{\PYGZdl{}}\PYG{n}{roleClasses}\PYG{+w}{ }\PYG{o}{:=}
\PYG{+w}{  }\PYG{p}{//}\PYG{n+nt}{RoleClass}\PYG{p}{[}\PYG{n+na}{@RefBaseClassPath}\PYG{o}{=}\PYG{l+s+s1}{\PYGZsq{}PlantEquipment\PYGZsq{}}\PYG{p}{]}
\PYG{k}{for}\PYG{+w}{ }\PYG{n+nv}{\PYGZdl{}}\PYG{n}{ie}\PYG{+w}{ }\PYG{o+ow}{in}\PYG{+w}{ }\PYG{p}{//}\PYG{n+nt}{InternalElement}
\PYG{+w}{ }\PYG{k}{where}\PYG{+w}{ }\PYG{k}{some}\PYG{+w}{ }\PYG{n+nv}{\PYGZdl{}}\PYG{n}{rc}\PYG{+w}{ }\PYG{o+ow}{in}\PYG{+w}{ }\PYG{n+nv}{\PYGZdl{}}\PYG{n}{roleClasses}\PYG{+w}{ }\PYG{k}{satisfies}
\PYG{+w}{  }\PYG{n+nv}{\PYGZdl{}}\PYG{n}{ie}\PYG{p}{/}\PYG{n+nt}{RoleRequirements}\PYG{p}{/}\PYG{n+na}{@RefBaseRoleClassPath}\PYG{+w}{ }\PYG{o}{=}
\PYG{+w}{   }\PYG{n+nf}{string\PYGZhy{}join}\PYG{p}{(}\PYG{n+nv}{\PYGZdl{}}\PYG{n}{rc}\PYG{p}{/}\PYG{k}{ancestor\PYGZhy{}or\PYGZhy{}self}\PYG{p}{::}\PYG{n+nt}{*}\PYG{p}{/}\PYG{n+na}{@Name}\PYG{p}{,}\PYG{l+s+s1}{\PYGZsq{}/\PYGZsq{}}\PYG{p}{)}
\PYG{+w}{ }\PYG{k}{return}\PYG{+w}{ }\PYG{n+nv}{\PYGZdl{}}\PYG{n}{ie}
\end{Verbatim}
\textbf{SPARQL:}
\begin{Verbatim}[commandchars=\\\{\}]
\PYG{k}{SELECT} \PYG{n+nv}{?ie}
\PYG{k}{WHERE} \PYG{p}{\PYGZob{}}
  \PYG{n+nv}{?ie} \PYG{n+nn}{rdf}\PYG{p}{:}\PYG{n+nt}{type} \PYG{n+nn}{aml}\PYG{p}{:}\PYG{n+nt}{InternalElement} \PYG{p}{.}
  \PYG{n+nv}{?ie} \PYG{n+nn}{aml}\PYG{p}{:}\PYG{n+nt}{hasRoleRequirement}\PYG{o}{/}\PYG{p}{(}\PYG{n+nn}{rdfs}\PYG{p}{:}\PYG{n+nt}{subClassOf}\PYG{p}{)}\PYG{o}{*}
      \PYG{n+nn}{aml}\PYG{p}{:}\PYG{n+nt}{PlantEquipment} \PYG{p}{.\PYGZcb{}}
\end{Verbatim}
\end{listing}
Both xPath and xQuery operate directly on the XML-tree, and have therefore no mechanisms to handle ReferencePaths. 
xQuery offers some mechanisms to circumvent this limitation:
Technically, the second scenario could still be solved using xQuery. 
The approach includes reconstructing the reference path of all RoleClasses, iterating over them, and individually comparing them to the desired value (s. Listing~\ref{list:QueriesScenario23}). 
However, this approach is very limited in scope, e.g. capturing the open-ended transitive closure statement from Query 3 in this way is not possible. 
Using SPARQL, due to the support of Property Paths, all three queries can be answered in a much more compact and arguably more intuitive form (s. Listing~\ref{list:QueriesScenario23}).

\subsubsection*{Use Case 2: Automatic Model Creation}
In the second use case, a more complex query will be formulated, with the aim of automatically producing a Material Flow Graph of the Four-Tank-Mixing System. The automatic generation of material flow graphs or other causal graphs has many applications in areas like the automatic generation of simulation models\cite{Sierla.08.09.202011.09.2020}\cite{Ramonat}, or the contextualization and diagnosis of data anomalies\cite{Ramonat.2024} and alarm floods\cite{Arroyo.2015}\cite{Kunze.2023}\cite{VogelHeuser2024}. 

As in the previous Use Case, two scenarios with different levels of complexity will be covered: 

\begin{enumerate}
    \item \textit{Scenario 1 - Bidirectional FlowGraph}:
    
    Create a flow graph assuming bidirectional flow. 
        
    \item \textit{Scenario 2 - Valve-state dependent FlowGraph}:
    
    Create a valve-state dependent directional flow graph. 
\end{enumerate}

In the first scenario, the content of the AutomationML-file should be used to create a material flow graph of all possible paths through the system. In the second scenario, the query was extended to include information about the state of valves, as they enable or disable product flow in parts of the system. 

Therefore, both queries build upon the same basic graph pattern, but extend it with additional conditions that have to be met.

In this use case, it is assumed that the material flow interface in AutomationML was modeled using the InterfaceClass \emph{Port} or interfaces that inherit from this class. The \emph{Port}-InferfaceClass is defined in the AutomationMLBaseInterfaceLibrary. 
In addition, any valves in the model can have a boolean ValveState-Attribute that indicates, whether the valve is open or closed. 

\begin{figure}[ht]
\includegraphics[width=\linewidth]{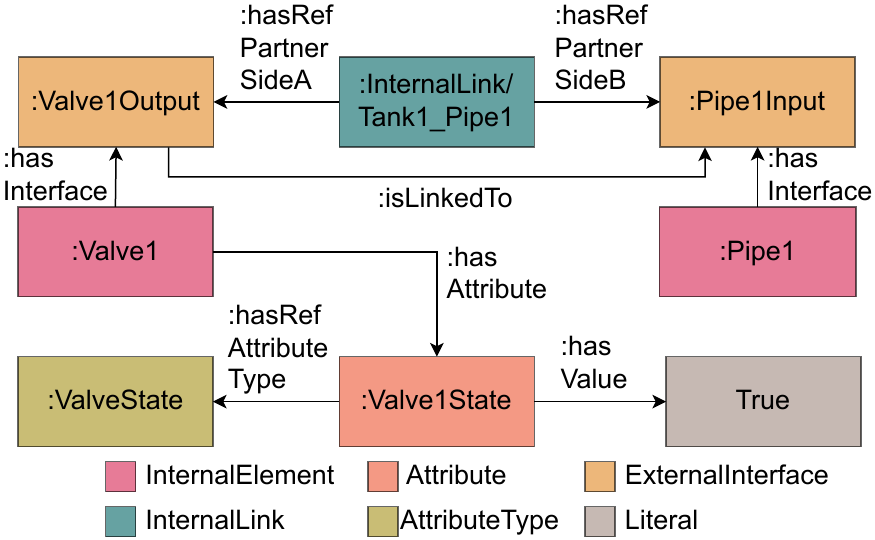}
\caption{Example of a valve that is connected to a pipe. The valve has a state, that defines whether it is open or not. }
\label{fig:InternalLinkFlowGraph}
\end{figure}

Since it does not support multi-step-patterns, xPath was unable to create a material flow graph for either of the three scenarios. 
Using xQuery, a Flow Graph for the first Scenario could be created. However, adding additional checks and constraints quickly results in very deeply nested loops over all possible connections. 
While technically possible, the queries this approach produced were opaque and error-prone. 

SPARQL, on the other hand, offers the possibility to define repeating patterns inside the graph that should be traversed. 
Due to these features, all queries can be answered using a more straightforward and compact SPARQL-Query. 

A SPARQL query that solves the second scenario can be found in Listing~\ref{list:FlowGraphQuery}.

\begin{listing}[ht]
\caption{\sffamily\fontencoding{T1}\fontseries{m}\fontsize{7}{8.4}\selectfont SPARQL query, that creates a material flow graph while considering valve states.}
\label{list:FlowGraphQuery}
\begin{Verbatim}[commandchars=\\\{\}]
\PYG{k}{construct}\PYG{p}{\PYGZob{}}\PYG{n+nv}{?s} \PYG{n+nn}{aml}\PYG{p}{:}\PYG{n+nt}{flows} \PYG{n+nv}{?o}\PYG{p}{.\PYGZcb{}} \PYG{k}{where} \PYG{p}{\PYGZob{}}
    \PYG{n+nv}{?s} \PYG{k}{a} \PYG{n+nn}{aml}\PYG{p}{:}\PYG{n+nt}{InternalElement}\PYG{p}{.}
    \PYG{n+nv}{?o} \PYG{k}{a} \PYG{n+nn}{aml}\PYG{p}{:}\PYG{n+nt}{InternalElement}\PYG{p}{.}
    \PYG{n+nv}{?s}   \PYG{n+nn}{aml}\PYG{p}{:}\PYG{n+nt}{hasInterface} \PYG{n+nv}{?i1}\PYG{p}{.}
    \PYG{n+nv}{?o}   \PYG{n+nn}{aml}\PYG{p}{:}\PYG{n+nt}{hasInterface} \PYG{n+nv}{?i2}\PYG{p}{.}
    \PYG{n+nv}{?i1}  \PYG{n+nn}{aml}\PYG{p}{:}\PYG{n+nt}{isLinkedTo}   \PYG{n+nv}{?i2}\PYG{p}{.}
    \PYG{k}{OPTIONAL} \PYG{p}{\PYGZob{}}
      \PYG{n+nv}{?s}   \PYG{n+nn}{aml}\PYG{p}{:}\PYG{n+nt}{hasAttribute} \PYG{n+nv}{?vs}\PYG{p}{.}
      \PYG{n+nv}{?vs}  \PYG{n+nn}{aml}\PYG{p}{:}\PYG{n+nt}{hasRefAttributeType} \PYG{n+nn}{aml}\PYG{p}{:}\PYG{n+nt}{ValveState}\PYG{p}{.}
      \PYG{n+nv}{?vs}  \PYG{n+nn}{aml}\PYG{p}{:}\PYG{n+nt}{hasAttributeValue} \PYG{n+nv}{?state}\PYG{p}{.}
    \PYG{p}{\PYGZcb{}}
    \PYG{k}{Filter}\PYG{p}{(}\PYG{o}{!}\PYG{n+nf}{bound}\PYG{p}{(}\PYG{n+nv}{?state}\PYG{p}{)} \PYG{o}{||} \PYG{n+nv}{?state} \PYG{o}{=} \PYG{l+s}{\PYGZsq{}true\PYGZsq{}}\PYG{p}{).}
\PYG{p}{\PYGZcb{}}
\end{Verbatim}
\end{listing}

\begin{figure}[htbp]
\includegraphics[width=\linewidth]{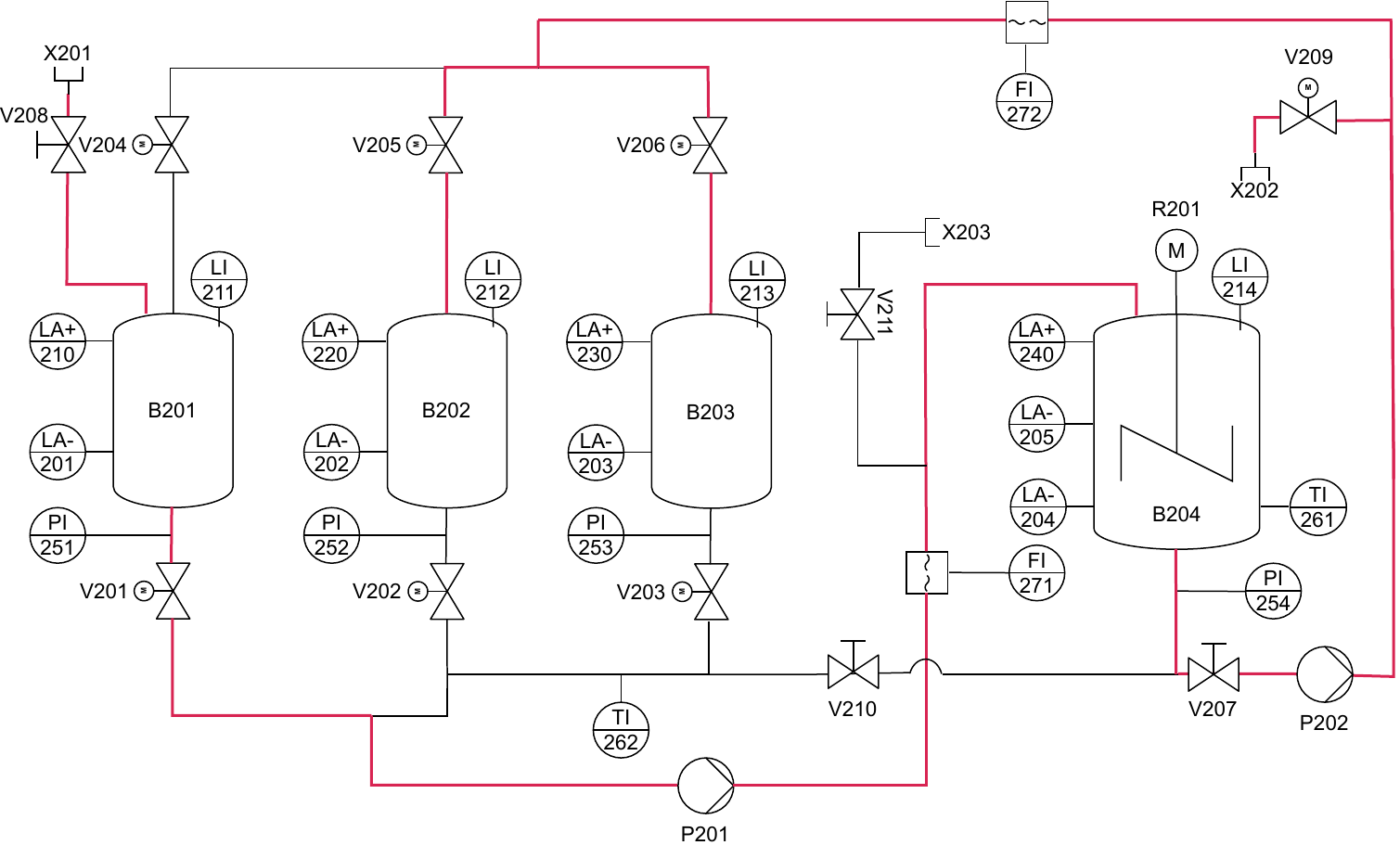}
\caption{Illustration of the mixing module's resulting flow graph (shown in red), if some of the valves are closed.}
\label{fig:FlowGraph}
\end{figure}

\subsubsection*{Comparison of xPath, xQuery, SPARQL}
The three query languages show differences in their ability to extract information from the AutomationML model. 
All three query languages are able to perform basic queries that encompass selecting elements based on filtering conditions. 
However, once the AutomationML-ReferencePaths become involved, difficulties emerge in xPath and xQuery.

For this comparison, we  refer to the different categories of filtering conditions from Section \ref{sec:complexGraphQueries}: 
In First Degree Filtering and Comparisons, the filter condition applies to the direct attributes that an element has. These tasks also encompass all queries in which only the ReferencePath is of interest, and the information at the destination of the path is irrelevant. All query languages were able to fulfill these tasks. 
In Second-Order Filtering, the filter condition applies to an element that the first element references via a reference path.
For these queries, xPath is unable to retrieve the targeted elements. xQuery, on the other hand, is technically able to answer those queries through a combination of loops and string reconstructions and comparisons. However, the resulting queries are complicated and not robust. 

In n-th order filtering, the filter condition applies to an element that the first element references through a chain of two or more subsequent reference paths. This case also includes use cases like multilevel inheritance or complex path patterns. In these cases, xQuery is not suitable to create functional queries. 
All of these shortcomings of xPath and xQuery to query AutomationML files stem from the mismatch between the XML tree structure that both languages were designed for, and the labeled directed multigraph that the contents of an AutomationML file actually represent.
Since it is a dedicated graph query language, SPARQL is able to handle such use cases, while also offering explicit support for more advanced and graph specific features like property paths and complex graph patterns. 

In summary, the comparison between the query languages shows that xPath and xQuery are at best sufficient for simple queries and information retrieval. If the scenarios become more complex, SPARQL offers greater functionalities and ease of use.

\subsection{Model Validation}
Besides the potential for querying, representing an AutomationML file in the form of an ontology also offers possibilities for validation of the file's content. 
A case study was designed to compare three common validation languages - XSD, Schematron and SHACL - for their suitability for AutomationML data. 

XML Schema, usually abbreviated XSD, operates directly on the XML files. It is commonly used to validate the syntactic correctness of an XML file. 
Schematron is an XML schema language used for rule-based validation of XML documents. In comparison to XSD, Schematron offers a more flexible, rule-based approach.
SHACL on the other hand is a language for validating RDF data. It provides mechanisms to define constraints on RDF graphs, ensuring that data conforms to specific shapes and patterns. 

\subsubsection*{Use Case: Library-Level Rules}
In this use case, the validation concerns the correct usage of elements that were imported from user defined libraries. 
These libraries are commonly accompanied by textual specifications that define the proper usage of the imported elements. 
As those specifications lack a formal or machine interpretable definition, their correct usage needs to be manually checked by domain experts. 
For this use case, the \textit{AutomationML Application Recommendation Automation Project Configuration} (AR APC) was considered. It is a framework that describes guidelines for the configuration of automation projects.\cite{ARAPC}

\begin{figure}[ht]
\includegraphics[width=\linewidth]{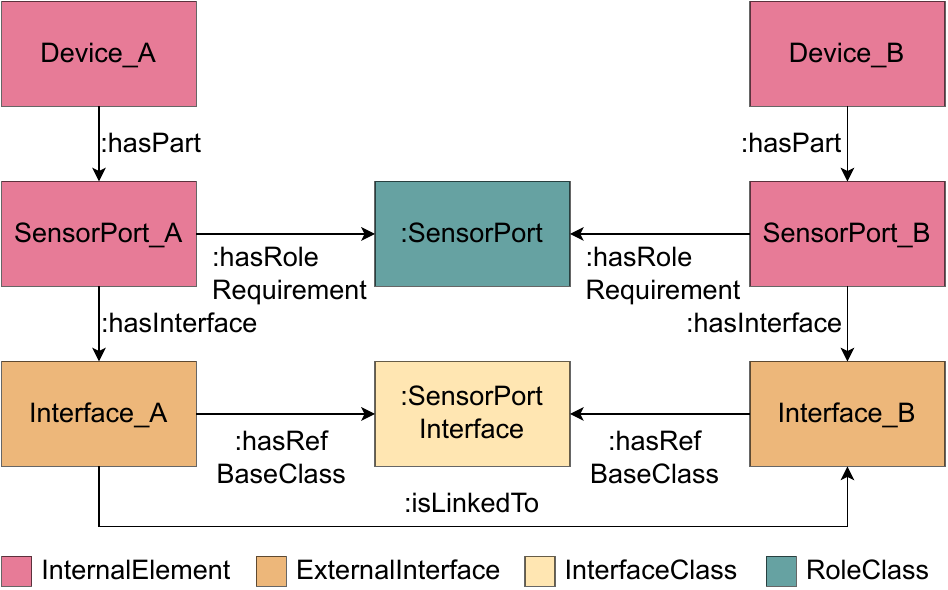}
\caption{Example of a valid connection of two Devices using the SensorPort element. }
\label{fig:AML2OWLSensorPortExample}
\end{figure}

For this example, the SensorPort-RoleClass was considered (s. Figure~\ref{fig:AML2OWLSensorPortExample}). 
It is a RoleClass used to model the connection between devices for the transfer of sensor signals. As such, it can contain an Interface that connects to SensorPortsInterfaces. 
The AR APC defines two constraints on this class:

\begin{itemize}
    \item \textit{Scenario 1 - Cardinality Constraint:}
    
    All Instances of the class can have at most one SensorPortInterface. 
    \item \textit{Scenario 2 - Connection Constraint:}
    
    SensorPort-Interfaces can only be connected to other SensorPort-Interfaces.  
\end{itemize} 

\begin{listing}[ht]
\caption{\sffamily\fontencoding{T1}\fontseries{m}\fontsize{7}{8.4}\selectfont Schematron and SHACL constraints for Scenario 2.}
\label{list:ValidationScenario23}
\center{\textbf{Schematron:}}
\begin{Verbatim}[commandchars=\\\{\}]
\PYG{n+nt}{\PYGZlt{}rule}\PYG{+w}{ }\PYG{n+na}{context=}\PYG{l+s}{\PYGZdq{}caex:InternalElement[}
\PYG{l+s}{ caex:RoleRequirements[@RefBaseRoleClassPath=}
\PYG{l+s}{ \PYGZsq{}ArApcRcLib/SensorPort\PYGZsq{}]]\PYGZdq{}}\PYG{n+nt}{\PYGZgt{}}
\PYG{+w}{ }\PYG{n+nt}{\PYGZlt{}assert}\PYG{+w}{ }\PYG{n+na}{test=}\PYG{l+s}{\PYGZdq{}}
\PYG{l+s}{  every \PYGZdl{}link in caex:InternalLink satisfies}
\PYG{l+s}{   (//caex:ExternalInterface[}
\PYG{l+s}{    @ID=\PYGZdl{}link/@RefPartnerSideA]/@RefBaseClassPath=}
\PYG{l+s}{     \PYGZsq{}ArApcIcLib/SensorPortInterface\PYGZsq{} and}
\PYG{l+s}{    //caex:ExternalInterface[}
\PYG{l+s}{    @ID=\PYGZdl{}link/@RefPartnerSideB]/@RefBaseClassPath=}
\PYG{l+s}{    \PYGZsq{}ArApcIcLib/SensorPortInterface\PYGZsq{})\PYGZdq{}}\PYG{n+nt}{\PYGZgt{}}
\PYG{+w}{ }\PYG{n+nt}{\PYGZlt{}/assert\PYGZgt{}}
\PYG{n+nt}{\PYGZlt{}/rule\PYGZgt{}}
\end{Verbatim}
\textbf{SHACL:}
\begin{Verbatim}[commandchars=\\\{\}]
\PYG{p}{:}\PYG{n+nt}{SPIConnectionShape}
  \PYG{k}{a} \PYG{n+nn}{sh}\PYG{p}{:}\PYG{n+nt}{NodeShape} \PYG{p}{;}
  \PYG{n+nn}{sh}\PYG{p}{:}\PYG{n+nt}{targetClass} \PYG{n+nn}{aml}\PYG{p}{:}\PYG{n+nt}{SensorPortInterface}\PYG{p}{;}
  \PYG{n+nn}{sh}\PYG{p}{:}\PYG{n+nt}{property} \PYG{p}{[}
    \PYG{n+nn}{sh}\PYG{p}{:}\PYG{n+nt}{path} \PYG{n+nn}{aml}\PYG{p}{:}\PYG{n+nt}{isLinkedTo}\PYG{p}{;}
    \PYG{n+nn}{sh}\PYG{p}{:}\PYG{n+nt}{class} \PYG{n+nn}{aml}\PYG{p}{:}\PYG{n+nt}{SensorPortInterface}\PYG{p}{;]} \PYG{p}{.}
\end{Verbatim}
\end{listing}

\subsubsection*{Comparison of XSD, Schematron and SHACL}
Both scenarios constitute conditional constraints, i.e. they check the value of an XML-Attribute (RefBaseClassPath) and only apply the validation if it matches a certain value (\textit{CommunicationRoleClassLib/PhysicalDevice/SensorPort}). 

While XSD can validate basic data types and structure, it doesn't support conditional validation and thus lacks the ability to handle complex relationships between elements or logic.
Compared to XSD, Schematron and SHACL are far more expressive and flexible. 
However, Schematron is tied to XML’s tree structure and can't natively handle ReferencePaths or InternalLinks. Solving validation scenarios thus involves complex iterations over the data, just as it was the case with xQuery.

Between the three approaches, XSD is most suitable for basic structural validation (e.g., ensuring elements appear in the correct order or enforcing data types), while both Schematron and SHACL can be used for cross-element validation with a lower level of complexity (s. Table~\ref{tab:ValidationLanguages}). 
Compared to the other validation approaches, SHACL has access to a semantically described relationship model and can therefore define complex graph shapes that the data must conform to.
As the complexity of the validation pattern increases, SHACL's advantages are expected to grow more pronounced. 

\begin{table}[htbp]
\caption{Capabilities of evaluated Validation Languages}
\label{tab:ValidationLanguages}
\setlength{\tabcolsep}{3pt}
\begin{tabular}{|p{173pt}|c|c|c|}
\hline
Validation Property                            & \hspace{8pt} XSD \hspace{8pt} & Schematron &  \hspace{5pt} SHACL \hspace{5pt} \\
\hline 
Basic Structural Validation                & +     & +      & +      \\
Library-Level Cardinality Validation      & -     & +      & +      \\
Library-Level Connection Constraint       & -     & o      & +      \\
\hline
\end{tabular}
\label{tab:ValidationFunctionalities}
\end{table}

\section{Conclusion}\label{sec:conclusion}
AutomationML is a widespread and expressive exchange format in the automation domain.
In the past, multiple attempts have been made to incorporate the knowledge contained in AutomationML files into knowledge graphs.
While these approaches have been partially successful, no common method of mapping the information contained in AutomationML-files to knowledge graphs exists. 
To mitigate this issue, this article provides:
\begin{enumerate}
    \item An AutomationML ontology that is compatible with version 3 of the CAEX specification.
    \item A declarative mapping that offers practitioners a straightforward transformation of AutomationML files to RDF triples.
\end{enumerate}
The ontology\footnote{https://github.com/hsu-aut/IndustrialStandard-ODP-AutomationML} as well as the mapping files and applications\footnote{https://github.com/hsu-aut/aml2owl} are freely available to practitioners and can be found on Github.
They were validated on multiple exemplary AutomationML files and were able to capture and adequately replicate the information contained in the file. 

Two studies were conducted to evaluate the mappings' potential for graph queries and validation. They showed that the mapping to RDF offers expanded possibilities for querying as well as data validation.

\subsection{Discussion}
In comparison to previous approaches to mapping AutomationML files to RDF graphs, the use of an RML mapping offers multiple advantages. 
For one, all used artifacts (RML mapping and SPARQL queries) are independent of any implementation language. 
In case of updates to AutomationML specification, only the mapping needs to be updated, while the tools that use them can remain the same. This should result in lower total maintenance. 
The evaluation showed that mapping AutomationML files to RDF enables more complex use cases for querying and validation.
Since common XML-based solutions were designed to handle a different type of graph structure, they can only operate on a subset of the information that AutomationML actually provides.  

The experiments regarding automatic validation show that mapping the AutomationML file to RDF and subsequently applying SHACL validation rules improve upon the validation capabilities of the current approach using XSD. 
The same conclusion was reached regarding the querying capabilities, in which SPARQL proved to be more capable and user friendly than xPath and xQuery. 

\subsection{Outlook and Future Work}
This work is part of an ongoing effort to create a formal description of AutomationML. 
A plug-in for the AutomationML Editor is currently in development. It can be used to translate an AutomationML file to RDF and then query it using SPARQL or evaluate it against SHACL validation rules. 

The artifacts provided by this work enable new uses of the information in an AutomationML file. 
The capability to formulate complex queries and create models based on the file contents allow new applications in the field of informed machine learning, in which prior knowledge of a system is used to improve various common AI tasks such as anomaly detection or diagnosis. These approaches would especially benefit from a connection between the static information from the AutomationML file with run time data from the system. Additionally, having a simple way to represent the files contents in an RDF graph allows the integration of the knowledge from the AutomationML file with other information artifacts about a system. 

\section*{Acknowledgment}
We would like to express our gratitude to the members of the FD4AML working group from the AutomationML association for their valuable feedback and constructive criticism, which significantly contributed to the improvement of this work. In particular, we thank Miriam Schleipen, Tina Mersch, Sandra Zimmermann and Aljosha Köcher for their insightful comments and suggestions that helped refine the analysis and presentation of our findings.

\bibliographystyle{IEEEtranIES}
\bibliography{main.bib}

\end{document}